\documentclass{article}

\usepackage{microtype}
\usepackage{graphicx}
\usepackage{subfigure}
\usepackage{booktabs} 
\usepackage{amsfonts}
\usepackage{amsmath}
\usepackage{dsfont}

\usepackage{hyperref}

\usepackage[accepted]{style/icml2021}

\begin{document}

\twocolumn[
\icmltitle{Segmenting Hybrid Trajectories using Latent ODEs}

\begin{icmlauthorlist}
\icmlauthor{Ruian Shi}{to,vec}
\icmlauthor{Quaid Morris}{vec,msk}
\end{icmlauthorlist}

\icmlaffiliation{to}{University of Toronto}
\icmlaffiliation{vec}{Vector Institute, Toronto}
\icmlaffiliation{msk}{Memorial Sloan Kettering Cancer Center}

\icmlcorrespondingauthor{Ruian Shi}{ian.shi@mail.utoronto.ca}

\icmlkeywords{LatSegODE, Time Series Segmentation, Neural ODE, Latent ODE, Machine Learning}

\vskip 0.3in
]

\printAffiliationsAndNotice{}

\begin{abstract}
    Smooth dynamics interrupted by discontinuities are known as hybrid systems and arise commonly in nature. Latent ODEs allow for powerful representation of irregularly sampled time series but are not designed to capture trajectories arising from hybrid systems. Here, we propose the Latent Segmented ODE (LatSegODE), which uses Latent ODEs to perform reconstruction and changepoint detection within hybrid trajectories featuring jump discontinuities and switching dynamical modes. Where it is possible to train a Latent ODE on the smooth dynamical flows between discontinuities, we apply the pruned exact linear time (PELT) algorithm to detect changepoints where latent dynamics restart, thereby maximizing the joint probability of a piece-wise continuous latent dynamical representation. We propose usage of the marginal likelihood as a score function for PELT, circumventing the need for model-complexity-based penalization. The LatSegODE outperforms baselines in reconstructive and segmentation tasks including synthetic data sets of sine waves, Lotka Volterra dynamics, and UCI Character Trajectories.
\end{abstract}

\section{Introduction}
\label{intro}
\begin{figure}[ht]
    \centering
    \includegraphics[width=0.47\textwidth]{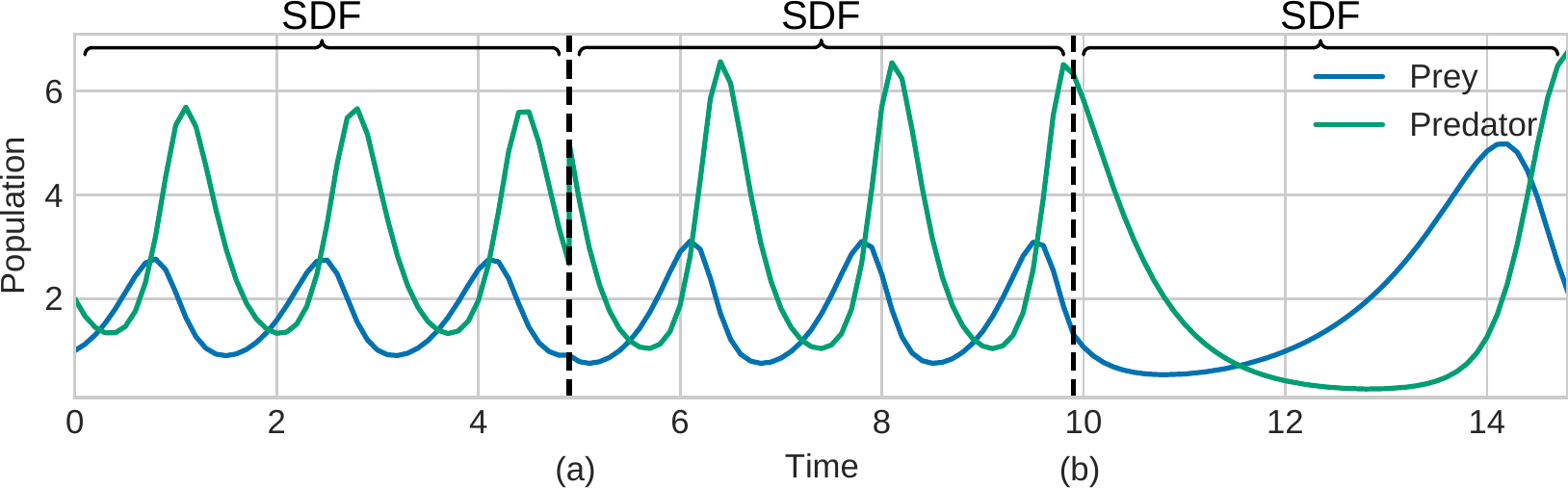}
    \caption{A Lotka-Volterra hybrid trajectory composed of three smooth dynamical flows. The plot shows populations of predators and prey over time. At time (a), a jump discontinuity occurs. At time (b), a distributional shift in dynamical coefficients occurs.}
    \label{fig:lv_example}
\end{figure}

The complexity of modelling time-series data increases when accounting for discontinuous changes in dynamical behavior. As a motivational example, consider the Lotka-Volterra equations, a simplified model of predator-prey interactions. The system is described by the pair of ordinary differential equations (ODEs):
\begin{equation}
    \frac{dx}{dt} = \alpha x - \beta x y \hspace{2em} \frac{dy}{dt} = \delta x y - \gamma y
\end{equation}
where $x$ and $y$ are the population size of predators and prey respectively. Coefficients $\alpha, \beta, \delta, \gamma$ describe interaction characteristics, such as the rate of encounter, and rate of successful predation per encounter. When these parameters are fixed, modelling this system from observed population trajectories is straightforward. However, external factors can perturb the system. Additional predators can suddenly be introduced via migration midway in an observed population trajectory, causing a jump discontinuity in the trajectory. The coefficients describing predator-prey interaction may also abruptly change, instantaneously changing the dynamical mode of the system. Systems featuring smooth dynamical flows (SDFs) interrupted by discontinuities are known as hybrid systems \cite{van2000introduction}. These discontinuities can arise as discrete jumps or instantaneous switches in dynamical mode \cite{ackerson1970state}, shown in Figure \ref{fig:lv_example} at times (a) and (b) respectively. We propose a method to model the hybrid trajectories which arise from hybrid systems.

Recently, the Latent ODE architecture \cite{rubanova2019latent} has been introduced to represent time series using latent dynamical trajectories. However, Latent ODEs are not designed to model discontinuous latent dynamics and, thus,  represent hybrid trajectories poorly. Here, we propose the Latent Segmented ODE (LatSegODE), an extension of a Latent ODE explicitly designed for hybrid trajectories. Given a base model Latent ODE trained on the segments of SDFs between discontinuities, we apply the Pruned Exact Linear Time (PELT) search algorithm \cite{killick2012optimal} to model hybrid trajectories as a sequence of samples from the base model, each with a different initial state. The LatSegODE detects the positions where the latent ODE dynamics are restarted with a new initial state, thus modelling hybrid trajectories using a piece-wise continuous latent trajectory. We provide a novel way to use deep architectures in conjunction with offline changepoint detection (CPD) methods. Using the marginal likelihood under the Latent ODE as a score function, we find the Bayesian Occam's Razor \cite{mackay1992bayesian} effect automatically prevents over-segmentation in CPD methods. 

We evaluate LatSegODE on data sets of 1D sine wave hybrid trajectories, Lotka-Volterra hybrid trajectories, and a synthetically composed UCI Character Trajectories data set. We demonstrate that the LatSegODE interpolates, extrapolates, and finds the changepoints in hybrid trajectories with high accuracy compared to current baseline methods. 

\section{Background}
\subsection{Latent ODEs}
The Latent ODE architecture \cite{rubanova2019latent} is an extension of the Neural ODE method \cite{chen2018neural}, which provides memory-efficient gradient computation without back-propagation through ODE solve operations. Neural ODEs represent trajectories as the solution to the initial value problem: 
\begin{align}
    \frac{d h(t)}{dt} &= f_\theta(h(t), t) \\
    h_{0:N} &= \text{ODESolve}(f_\theta, h0, t_{0:N})
\end{align}
where $f_\theta$ is parameterized by a neural network, and $h(t)$ represents hidden dynamics. The continuous dynamical representation allows Neural ODEs to natively incorporate irregularly sampled time series.

Latent ODEs arrange Neural ODEs in an encoder-decoder architecture. Observed trajectories are encoded using a GRU-ODE architecture \cite{de2019gru, rubanova2019latent}. The GRU-ODE combines a Neural ODE with a gated recurrent unit (GRU) \cite{cho2014learning}. Observed trajectories are encoded by the GRU into a hidden state, which is continuously evolved between observations by a Neural ODE parameterized by neural network $f_\theta$. The GRU-ODE encodes the observed data sequence into parameters for a variational posterior. Using the reparameterization trick \cite{kingma2013auto}, a differentiable sample of the latent initial state $z0$ is obtained. A Neural ODE parameterized by neural network $f_\Psi$ deterministically solves a latent trajectory from the latent initial state. Finally, a neural network $f_\Phi$  decodes the latent trajectory into data space. The Latent ODE architecture can thus be represented as:
\begin{align}
    \mu_{z0}, \sigma_{z0}^2 &= \text{GRUODE}_{f_\theta}(x_{1:N}, t_{1:N})\\
    z0 &\sim q(z0 | x_{1:N}) = \mathcal{N}(\mu_{z0}, \sigma_{z0}^2) \\
    z_{1:N} &= \text{ODESolve}(f_\Psi, z0, t_{1:N}) \\
    x_{i} &\sim \mathcal{N}(f_{\Phi}(z_{i}), \sigma^2) \hspace{1em} \text{for} \hspace{1em} i = 1, ..., N
\end{align}
where $\sigma^2$ is a fixed variance term. The Latent ODE is trained by maximizing the evidence lower-bound (ELBO). Letting $X = x_{1:N}$, the ELBO is:
\begin{equation}
    \mathbb{E}_{z0 \sim q(z0|X)}[\log p(X)] - \text{KL}\left[q(z0|X)\;||\;p(z0)\right]
\end{equation}
\subsection{Representational Limitations of the Neural ODE}
Latent ODEs use Neural ODEs to represent latent dynamics, and thus inherit their representational limitations. The accuracy of an ODE solver used by a Neural ODE depends on the smoothness of the solution; the local error of the solution can exceed ODE solver tolerances when a jump discontinuity occurs \cite{calvo2008}. At a jump, adaptive ODE solvers will continuously reduce step size in response to increased error, possibly until numerical underflow occurs. Even if integration is possible across the jump, it is slow, and the global error of the solution can be adversely affected \cite{calvo2003}. Typically, these issues can be easily avoided by restarting ODE solutions at the discontinuity but this requires these positions to be known. Classical methods use the increase in local error or adaptive rejections associated with jump discontinuity as criteria to restart solutions \cite{calvo2008}. Recently, Neural Event ODEs \cite{chen2020learning} uses a similar paradigm of discontinuity detection, using an event function parameterized by a neural network to detect locations to restart the ODE solution. With all event detection approaches, failure to accurately detect jump discontinuity will cause the local error bound decrease to a lower order \cite{stewart2011}. Hybrid trajectories with discontinuous change in the dynamical coefficients present different but still hard modeling challenges due to the representational limitations of Neural ODEs. 

Latent ODEs do not circumvent these limitations, and cannot generalize in hybrid trajectories. When a hybrid trajectory is encountered, the Latent ODE can only encode the exact sequence of SDFs into a single latent representation. Should a permutation of these SDFs arise at test time, the Latent ODE will not be able to reconstruct the test trajectory. 

\begin{figure*}[t]
    \centering
    \includegraphics[width=0.9\textwidth]{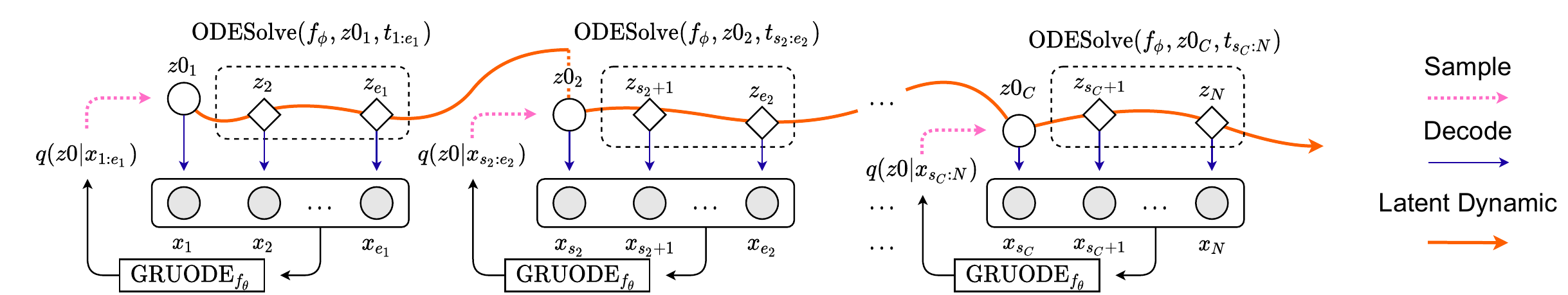}
    \caption{Schematic of the LatSegODE reconstructing a hybrid trajectory. Arrows indicate computation flow. Data in each segment is encoded into parameters for the variational posterior, from which a latent initial state is sampled. Each latent segment is solved using shared latent dynamic $f_\Phi$, which continues until the next point of change. The latent trajectory is decoded into data space. At evaluation time, an arbitrary number of changepoints can be detected by the PELT algorithm. Plot adapted from \cite{rubanova2019latent}.}
    \label{fig:schematic}
\end{figure*}

\section{Method}
The LatSegODE detects positions of jump discontinuity or switching dynamical mode by representing a hybrid trajectory as a piece-wise combination of samples from a learned base model Latent ODE. At each changepoint, the latent dynamics of the base model are restarted from a new initial state. We apply the PELT algorithm to efficiently search through all possible positions to restart ODE dynamics, and return changepoints that correspond to the positions of restart which maximize the joint probability of a hybrid trajectory. This avoids the need to train an event detector, and guarantees optimal segmentation, but the LatSegODE requires the availability of a training data set of SDFs on which the base model can be trained.

\subsection{Extension to Hybrid Trajectories}
We first define the class of hybrid trajectories which can be represented by the LatSegODE. Consider a sequential series of data $X = x_1, x_2, ..., x_N$ and associated times of observation $T = t_1, t_2, ..., t_N$. We represent a hybrid trajectory as a piece-wise sequence of $C$ continuous dynamical segments. Each observed data point can only belong to a single segment. Each segment is bounded by starting index $s_i$ and ending index $e_i$, where $0 \leq  i \leq C$,  $s_0 = 1$, and $e_C = N$. Segments are sequential and do not intersect, i.e.,  $s_{i+1} = e_{i} + 1$. The boundaries of segments represent locations of jump discontinuity or switch in dynamical mode. The trajectory within each segment is represented by a sample from the base model Latent ODE.

The LatSegODE can be applied to hybrid trajectories containing an unknown number and order of SDFs. The LatSegODE aims to approximate each SDF using a segment. Using offline CPD, the LatSegODE detects positions of jump discontinuity or switching dynamical mode, and introduces a latent discontinuity at the timepoint indexed by $s_i$. At these timepoints, indexed by $s_i$, the latent dynamics are restarted from a new latent initial condition $z0_i$, which is obtained from the Latent ODE encoder network acting on segment data points $x_{s_i:e_i}$. The latent dynamics are solved using the same latent Neural ODE parameterized by $f_\Phi$. We provide a schematic visualizing LatSegODE hybrid trajectory reconstruction in Figure \ref{fig:schematic}. The example hybrid trajectory is represented by a sequence of base model Latent ODE reconstructions, each starting from a new initial latent state which can discontinuously jump from the previous dynamic. An arbitrary number of restarts can be detected at test time.

To finish the problem formulation, we define $\mathcal{I}$ as the unknown ground truth set of segment boundaries and latent initial states, such that each hybrid trajectory is associated with set:
\begin{equation}
    \mathcal{I} = (s_i, e_i, z0_i) \hspace{4em} 0 \leq i \leq C
\end{equation} 
Where $Z = z_{0:C}$, the joint log probability of an observed hybrid trajectory can be represented as:
\begin{equation}
   \log p(X, Z | s_{1:C}, e_{1:C}) = \sum^C_{i=0} \log p(x_{s_i : e_i}, z0_{i})
\end{equation}
This formulation assumes independence between observations in separate segments, such that $x_{s_i:e_i} \perp (X \setminus x_{s_i:e_i})$. While this assumption can be limiting in trajectories with long term dependencies, it also allows for increased reconstruction performance in the absence of inter-segment dependency. In these situations, given a trajectory with two dynamical modes, allowing latent dynamics to completely restart at the time of modal change allows for a better representation. In comparison, methods which cannot account for shifts in latent dynamics will be forced to adopt an averaged representation between the two dynamical modes. This intuition is later demonstrated in the experimental section.

We note that the LatSegODE does not represent the location of changepoints using a random process. Since event detection is non-probabilistic, the method is not suitable for hybrid trajectories which self-excite or otherwise change dynamical mode past the observed trajectory.

\subsection{Optimal Segmentation}

Given this formulation of hybrid trajectories, the key challenge is finding the unknown set $\mathcal{I}$ which maximizes the joint probability of an observed hybrid trajectory. We propose application of optimized search algorithms from the field of offline changepoint detection (CPD) to recover locations of jump discontinuity and switches in dynamical mode, and consequently $\mathcal{I}$. Through complexity penalization, these search algorithms can automatically determine the optimal number and location of segments without prior specification.

Offline CPD methods attempt to discover changepoints which define segment boundaries. A combination of segments which reconstruct a trajectory is referred to as a segmentation. We allow each observed timepoint to be a potential changepoint. Thus, the space of all possible segmentations is formed by all combinations of an arbitrary number of changepoints. At either extremes, placing no changepoints or a changepoint at each time of observation are both valid segmentations. The space of all possible segmentations grows exponentially ($2^N$) with the number of observations ($N)$. 

The optimal partitioning method \cite{1381461} uses dynamical programming to search through this large space of solutions. Where $\mathcal{C}$ is a cost function, $m$ is the number of changepoints, and $\tau$ is a set of changepoints such that $\tau_0 = 0, \tau_{m+1} = n$, it minimizes
\begin{equation}
    \sum_{i=1}^{m+1} \mathcal{C}(x_{\tau_{i-1} + 1: \tau_{i}}) + \beta
\end{equation}
with respect to $\tau$ using dynamic programming. Of all possible segmentations up to data index $t$, we let $F(t)$ represent the one which results in the minimal cost. This result is memoized. For a new data index $s > t$, we can extend the optimal solution via recursion 
\begin{equation}
    F(s) = \min_t F(t) + \mathcal{C}(x_{(t+1):s}) + \beta
\end{equation}
Thus, we begin by solving for $F(1)$, and incrementally extend the solution until $F(N)$, at which point the optimal segmentation is returned. The memoization of previous optimal sub-solutions allows a quadratic runtime with respect to number of observations. The full algorithm is provided in Appendix A. The $\beta$ term penalizes over-segmentation, and typically scales with the number of parameters introduced by each additional changepoint. When a maximum likelihood cost function is used without a $\beta$ penalty, optimal partitioning degenerates by placing a changepoint at each possible index. The presence of $\beta$ enforces a trade-off between accuracy and model complexity. With an appropriate $\beta$, this formulation also conveniently recovers the segmentation with the minimized Bayesian Information Criterion (BIC) \cite{schwarz1978estimating} through minimization of equation (11).

Choice of $\beta$ is a key challenge in using CPD methods with deep architectures. It is not always clear how many effective parameters are introduced by each additional segment, though this number is upper bounded by the dimensionality of the latent initial state. Additionally, the theoretical assumptions required by the BIC are violated by neural network architectures \cite{watanabe2013widely}. The LatSegODE circumvents these challenges by using the marginal likelihood under the Latent ODE as the score function for each segment.

We compute a Monte Carlo estimate of the marginal likelihood by importance sampling using a variational approximation to the posterior over the initial state:
\begin{align}
    &\log p(x_{s:e}) = \log \int{p(x_{s:e}|z0) \, p(z0) \, \text{d}z0}\\
    &= \mathbb{E}_{z0\sim q(z0|x_{s:e})} \left[ p(x_{s:e}|z0) \frac{p(z0)}{q(z0|x_{s:e})} \right]\\
    &= \frac{1}{M}\sum_{j=1}^{M} \mathcal{N}(\overline{x_{s:e}} | x_{s:e}, \sigma^2) \frac{\mathcal{N}(z0_j|0, 1)}{\mathcal{N}(z0_j | \mu_{z0}, \sigma^2_{z0})}
\end{align}
where $\overline{x_{s:e}}$ is the output of the Latent ODE base model, $\mu_{z0}, \sigma^2_{z0}$ is obtained by the GRU-ODE encoder, and $z0_j$ is sampled as $\mathcal{N}(\mu_{z0}, \sigma^2_{z0})$. The variance $\sigma^2$ is fixed, and set to the same value used to compute the ELBO during training. We take $M$ samples for the Monte Carlo estimate.

Because we use the marginal likelihood, the complexity of the recovered segmentation is implicitly regularized by the Bayesian Occam's Razor \cite{mackay1992bayesian}. Reflecting this, in our experiments, we show that the penalization term $\beta$ can be set to $0$ without over-segmentation. Thus, we can simply set $\mathcal{C}$ in equation (11) to be the marginal likelihood computed by equation (15), and solve for the set of changepoints $\tau$ which maximize the joint probability of the entire trajectory using optimal partitioning (the original objective is a minimization, but this can trivially be switched to maximization).

The quadratic runtime of optimal partitioning can be reduced to between $\mathcal{O}(N)$ and $\mathcal{O}(N^2)$ through the pruned exact linear time (PELT) \cite{killick2012optimal} algorithm. Using an identical search algorithm, PELT introduces a pruning condition which allows removal of sub-solutions from consideration. Given the existence of $K$ such that for all changepoint indexes $s, t, T$ such that $t < s < T$:
\begin{equation}
    \mathcal{C}(x_{(t+1):s}) + \mathcal{C}(x_{(s+1):T}) + K \leq \mathcal{C}(x_{(t+1):T})
\end{equation}
Then if 
\begin{equation}
    F(t) + \mathcal{C}(x_{(t+1):s}) + K \geq F(s)
\end{equation}
we are able to discard the changepoint $t$ from future consideration, asymptotically reducing the number of operations required. Due to noise in the estimates of the score function, finding an analytic method to determine $K$ is an area for further research. If $K$ is set too low, sub-optimal solutions are recovered. In practice, this issue is not limiting, as setting $K$ to a sufficiently high value allows for near-optimal solutions at the cost of higher runtime. This trade-off is documented in Appendix B. 

The computation of $F(t)$, the optimal segmentation up to length $t$, and Monte Carlo estimate of the marginal likelihood can all be batch parallelized using GPU computation. An implementation is available at: \url{https://github.com/IanShi1996/LatentSegmentedODE}.

\subsection{When can I use this method?}
The LatSegODE requires a Latent ODE base model trained on a family of SDFs. We propose two scenarios where SDFs may be available. First, the LatSegODE is applicable when a training set of hybrid trajectories with labelled changepoints exists. In this case, given a training set of $N$ hybrid trajectories $X = (x^{(i)}, t^{(i)})_{i=1}^N$ each with $C$ labelled SDF boundaries $(s_{k}, e_{k})_{k=0}^C$, we treat each $x^{(i)}_{s_j:e_j}$ as an independent training trajectory, and train on the union of all SDFs. The LatSegODE can also be applied when physical simulation is available. In these scenarios, the base model can be trained on trajectories which are simulated in the range of dynamical modes which we expect in hybrid trajectories at test time. These two use cases are illustrated in the first two experiments.

\section{Related work}
\label{sec:related-works}
\textbf{Switching Dynamical Systems}:
Hybrid trajectories have previously been modelled as Switching Linear Dynamical Systems (SLDS). We provide a non-exhaustive summary of these methods. Typically, trajectories are represented by a Bayesian network containing a sequence of latent variables, from which observations are emitted. Latent variables are updated linearly, while a higher order of latent variable represents the current dynamical mode. Structured VAEs \cite{johnson2016composing} introduce a discrete latent variable to control dynamical mode, and use a VAE observation model. GPHSMMs \cite{nakamura2017segmenting} uses a Gaussian Process observation model within a hidden semi-Markov model. Kalman VAEs integrate a Kalman Filter with a VAE observation model \cite{fraccaro2017disentangled}. Models in this class are generally trained via an inference procedure \cite{dong2020collapsed}, while several are fully differentiable \cite{kipf2019compile}. These methods are unsupervised, requiring no training data with labelled changepoint locations.

In contrast, the LatSegODE requires a base model to be trained on SDFs. It does not model dependency between segments unlike methods such as rSLDS \cite{linderman2017bayesian}. At evaluation time, the LatSegODE operates without specification of the number of segments or dynamical modes. This is an advantage compared to previously discussed works, where performance is sensitive to these hyperparameters \cite{dong2020collapsed}.

The Neural Event ODE \cite{chen2020learning} is closely related to the LatSegODE. It represents observed dynamics using a Neural ODE and trains a neural network to detect the positions and update values of a switching dynamical system. The Neural Event ODE can be trained in an unsupervised fashion, without prior knowledge of change point locations in training data. When extrapolating past observed data, it is able to introduce additional change points, which the LatSegODE cannot model. However, the Neural Event ODE inherits the same limitations as the Neural ODE: it cannot model a data set which cannot be described by a single ODE function in data space. So, for example, two different dynamics cannot start from the same observed point. This issue is elaborated in Appendix C. The LatSegODE circumvents these limitation by modelling the data using an ODE in latent space.

\textbf{Offline Changepoint Detection}:
The LatSegODE closely relates to offline CPD, and we refer to \citet{truong2020selective} for an in-depth review. The LatSegODE leverages search algorithms from offline CPD, but represents the behavior within segments using a complex generative model, as opposed to a simple statistical cost function. The use of the Latent ODE allows for higher representational power and extrapolation/interpolation within segments. However, training data is required to fit the base model and, as such, its total runtime is significantly higher. Other methods have incorporated deep architectures with CPD search methods \cite{lee2018time}, but use a sliding window search with predefined window size, and use a feature distance metric to determine boundaries as opposed to the marginal likelihood used by LatSegODE.

\textbf{Miscellaneous}:
A distantly related class of methods classify individual observations into class labels, which can be seen as segmentation \cite{supratak2017deepsleepnet}. These approaches are distinct as they do not explicitly model dynamics, and require a fixed segment size and trajectory length, a limitation which the LatSegODE does not have. The LatSegODE does not treat positions of jump discontinuity or switching dynamical mode as a random variable, unlike methods that model these jumps as a random process \cite{mei2017neural, jia2019neural}.

\section{Experiments}
Here we investigate the LatSegODE's ability to simultaneously perform accurate reconstruction and segmentation on synthetic and semi-synthetic data sets. 

When training the base model, we mask observations from the last 20\% of the timepoints and 25\% of internal timepoints, this 25\% is shared across all training and test examples. When evaluating the model on the test set, we use the 55\% of unmasked timepoints to infer the initial states and perform segmentation, and then attempt to reconstruct the observations at the masked timepoints. We report the mean squared error (MSE) between ground truth and predicted observations on test trajectories. We benchmark against auto-regressive and vanilla Latent ODE baselines for reconstructive tasks. We augmented the input data for the vanilla Latent ODE with a binary-valued time series denoting changepoint positions. This ensures it has access to the same information as the LatSegODE. We report performance on an extrapolation region which assumes the last observed dynamical mode continues. We attempted to benchmark against Neural ODEs and Neural Event ODEs, but found that their training did not converge on any of our benchmarks (see Appendix C).

We benchmark the segmentation performance of the LatSegODE against classic CPD algorithms using Gaussian kernelized mean change \cite{arlot2019kernel}, auto regressive \cite{bai2000vector}, and Gaussian Process \cite{lavielle2006detection} cost functions. These are denoted RPT-RBF, RPT-AR, and RPT-NORM respectively. 

Segmentation performance is measured using the Rand Index \cite{rand1971objective}, the Hausdorff metric \cite{rockafellar2009variational}, and the F1 score. The Rand Index measures the overlap between the predicted segmentation and the ground truth segmentation. Given data points $x_{1:N}$, a membership matrix $A$ is defined such that $A_{ij} = 1$ if $x_i$ and $x_j$ are in the same segment. Otherwise, $A_{ij} = 0$. Membership matrices are generated for the ground truth segmentation $(A)$ and the predicted segmentation $(\Tilde{A})$. Using these two matrices, the Rand Index is calculated as:
\begin{equation}
    \frac{\sum_{i<j} \mathds{1}[A == \Tilde{A}]}{N(N-1) / 2}
\end{equation}
The Hausdorff metric is a measure of the maximal error between the predicted segmentation and the ground truth segmentation. Given a set of ground truth changepoints $\mathcal{T}$ and predicted changepoints $\mathcal{P}$, the Hausdorff metric is computed as:
\begin{equation}
     \max\left( \max_{\tau \in \mathcal{T}} \min_{\rho \in \mathcal{P}} |\tau - \rho|, \, \max_{\rho \in \mathcal{P}} \min_{\tau \in \mathcal{T}} |\tau - \rho|\right)
\end{equation}
We use the \verb+ruptures+ library \cite{truong2020selective} implementation of these baseline methods and metrics. 

We found that the segmentation baselines performed extremely poorly when using penalized detection of changepoints. In response, we simplified the problem for them by providing the correct number of changepoints, so that they only needed to choose the correct locations. In contrast, we did not provide LatSegODE with the number of changepoints, thus the evaluation was biased in favor of the baselines. Also, we excluded trajectories with zero changepoints from this benchmark because they are trivially correct. Irregular locations of data observation is handled by applying linear interpolation prior to segmentation. An extended description of baselines, metrics, and experimental set up is provided in Appendix D.

\subsection{Sine Wave Hybrid Trajectories}
We evaluate the LatSegODE on a benchmark data set of 1D sine wave hybrid trajectories. Here, we assume access to trajectories with labelled changepoint positions, one of the situations where the LatSegODE can be realistically applied. We generate 7500 hybrid trajectories each containing up to two changepoints. Between each changepoint, segment trajectories are sine waves generated under random parameters. We hold out $300$ validation trajectories, $150$ test trajectories, and train the LatSegODE base model on the SDFs contained in the remaining trajectories. Data parameters, model architecture and hyper-parameters are reported in Appendix E. In Figure \ref{fig:sine_reconstruct}, we provide a visual comparison of the LatSegODE against baselines on an example test set trajectory. 
\begin{figure}[ht]
    \centering
    \includegraphics[width=0.47\textwidth]{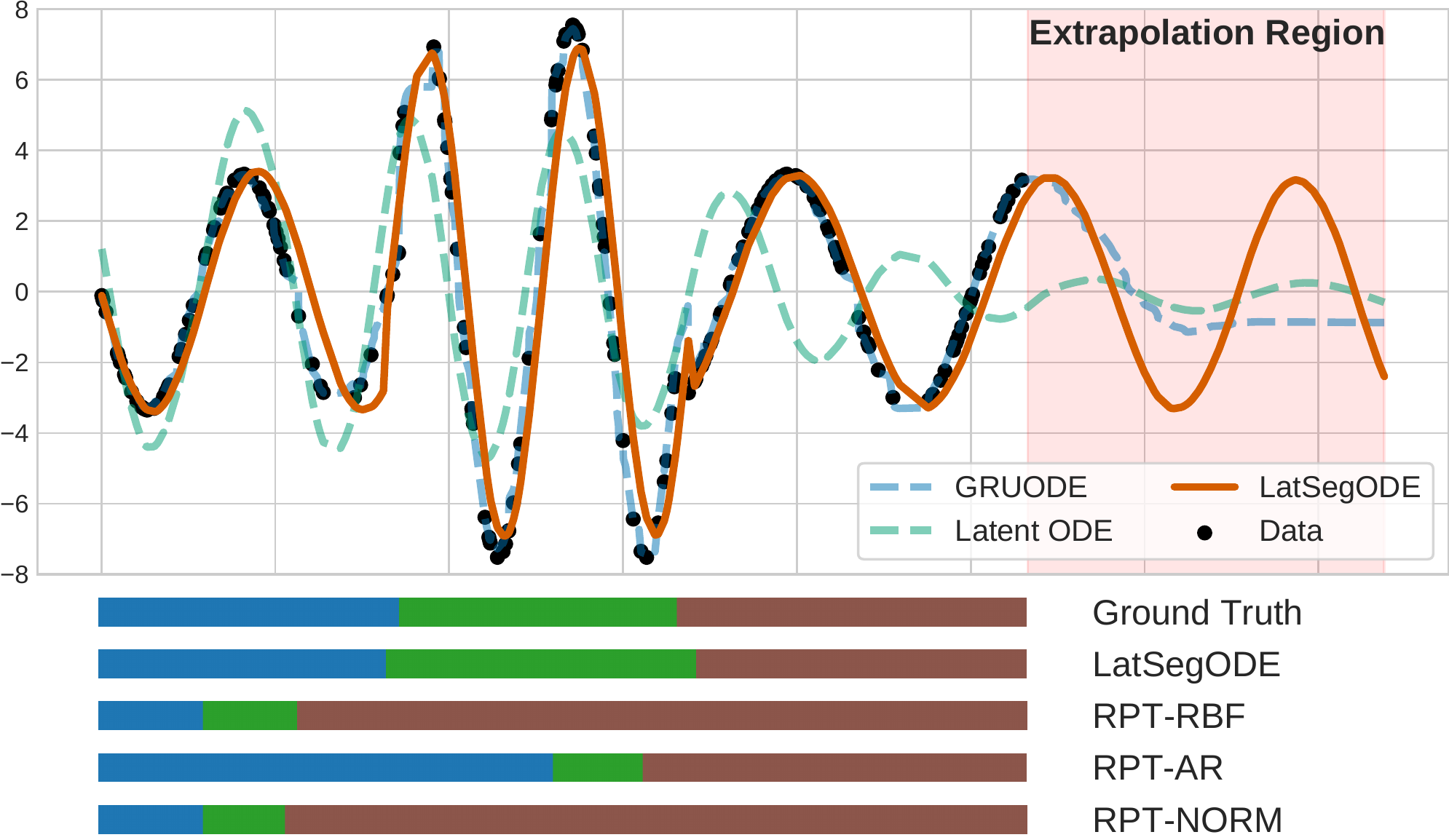}
    \caption{Comparison against baselines in a sample 1D Sine Wave hybrid trajectory. Top: Reconstructed trajectories are shown. Data in the extrapolation region is held out from all models during training. Bottom: Segmentation results are shown. Each distinctly colored region represents a segment.}
    \label{fig:sine_reconstruct}
\end{figure}
The LatSegODE outperforms baselines in both reconstruction and segmentation tasks. The presence of discontinuities prevent vanilla Latent ODEs from learning accurate representations. Although Latent ODEs can represent the initial SDFs, they lack the ability to represent switches to the dynamical mode. As time progresses, the Latent ODE reconstruction collapses near zero, a local minima which minimizes error given its reconstructive limitations. In contrast, because the LatSegODE can restart latent dynamics, it can represent trajectories with jump discontinuities. The LatSegODE provides an accurate reconstruction, and we see the periodic solution is cleanly captured in the extrapolation region. The GRU-ODE method can fit observed data well, but yields poor interpolations and extrapolations. The LatSegODE recovers the segmentation closest to the ground truth segmentation. The trends observed in this example trajectory are reflected in the overall test results, where the LatSegODE outperforms all baselines. These results are reported in Appendix F. We found that inclusion of the binary-valued changepoint location time series did not result in significant improvement, and we omit this feature from further experiments. We report the effects of the training set size and the number of samples per training trajectory on LatSegODE performance in Appendix K.
\begin{figure*}[ht]
    \centering
    \includegraphics[width=0.99\textwidth]{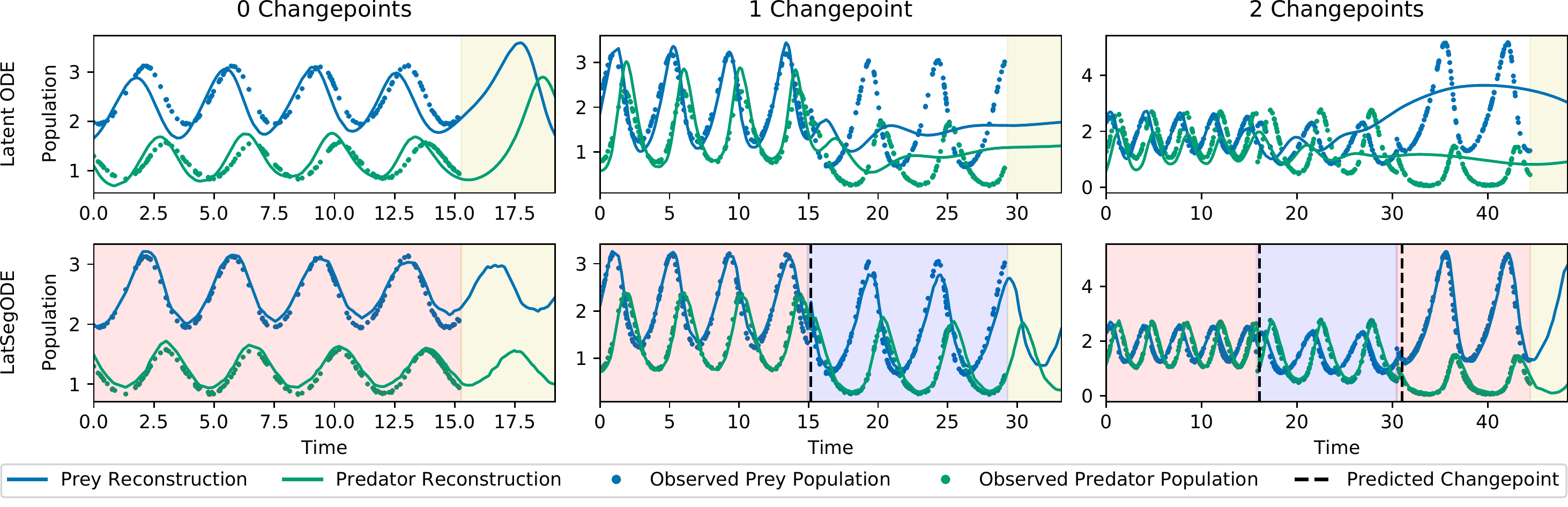}
    \caption{Comparison of reconstructions of Lotka Volterra hybrid trajectories. Top row contains baseline reconstruction by Latent ODE. Bottom row shows reconstruction by LatSegODE. Sample hybrid trajectories contain the same number of ground truth changepoints in each column. Ground truth segments are shown as a contiguous background color block. Yellow background indicates extrapolation region which assumes that the last observed dynamical mode continues. Visualization inspired by ruptures package \cite{truong2020selective}.}
    \label{fig:lv_results}
\end{figure*}
\subsection{Lotka-Volterra Hybrid Trajectories}
Next, we evaluate the LatSegODE on hybrid trajectories whose SDFs are the Lotka-Volterra dynamics described in equation (1). We simulate 34000/600/150 hybrid trajectories for the training/validation/test set. Lotka-Volterra dynamics are generated by randomly sampling coefficients $(\alpha, \beta, \delta, \gamma)$ from ranges $[(0.5, 1.5), (0.5, 1.5), (1.5, 2.5), (0.5, 1.5)]$ respectively. Each trajectory contains up to two changepoints, and at each changepoint we restart dynamics from new initial values sampled from $[(0.5, 1.5), (1.5, 2.5)]$. We re-sample the coefficient vector at changepoints, so the trajectories feature both jump discontinuity and switch of dynamical mode. We train the LatSegODE base model on the SDFs in the generated training trajectories. The vanilla Latent ODE baseline is trained on full hybrid trajectories, while other baselines were separately trained on both full trajectories and SDFs, with the best performing result reported. The data generation procedure, and model architectures/training is documented in Appendix G.

Results are reported in Table \ref{tab:lv-metrics}, where metrics are averages over 150 test trajectories. The LatSegODE outperforms baselines in both segmentation and reconstruction. An expanded evaluation with additional metrics and experiments is provided in Appendix H.

\begin{table}[ht]
\caption{Results on Lotka Volterra hybrid trajectories. Metrics generated using 150 test trajectories. Best result is bolded.}
\label{tab:lv-metrics}
\begin{center}
\begin{small}
\begin{sc}
\begin{tabular}{lcccr}
\toprule
Method & \multicolumn{1}{p{1cm}}{\centering Test\\MSE} & \multicolumn{1}{p{1cm}}{\centering Rand\\Index} & \multicolumn{1}{p{2cm}}{\centering Hausdorff\\Metric}\\
\midrule
LatSegODE    & \textbf{0.068} & \textbf{0.9464} & \textbf{47.67} \\
\midrule
GRU$\Delta t$ & 0.1718 & - & -\\
GRU-ODE    & 0.2747 & - & - \\
Latent ODE    & 0.6155 & - & -  \\
\midrule
RPT-RBF & - & 0.7956 & 84.7 \\
RPT-AR   & - & 0.6994 & 164.65 \\
RPT-NORM   & - & 0.7693 & 105.92  \\
\bottomrule
\end{tabular}
\end{sc}
\end{small}
\end{center}
\end{table}

\begin{figure*}[ht]
    \centering
    \includegraphics[width=0.99\textwidth]{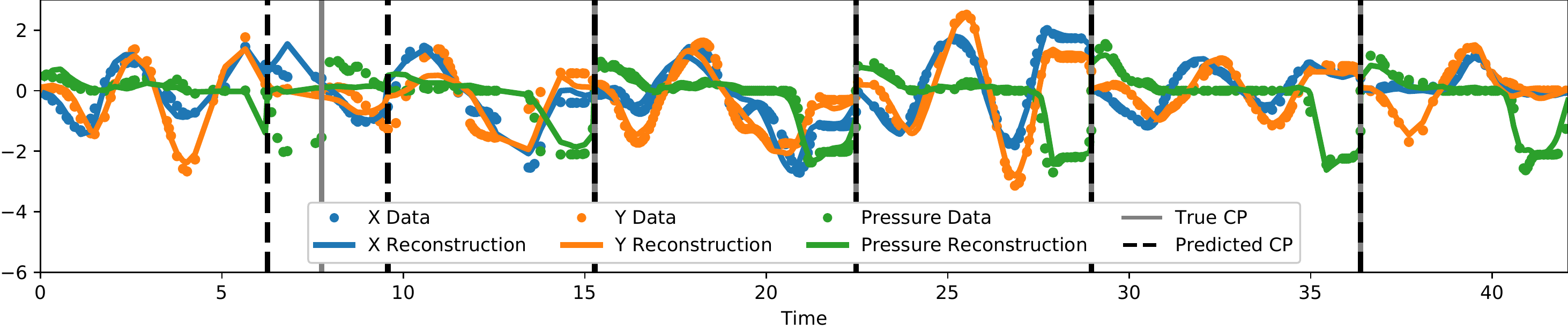}
    \caption{Example reconstruction on a long hybrid trajectory synthetically generated from UCI Character Trajectory data set. While most characters are accurately detected, an erroneous change point is introduced at $t \approx 6$. As segments are independent under PELT, future segments are not affected by this error, and reconstruction quality recovers after the introduction of the change point at $t \approx 9.5$.}
    \label{fig:ct}
\end{figure*}

In Figure \ref{fig:lv_results}, we show sample trajectory reconstructions from the LatSegODE versus the vanilla Latent ODE baseline. All vanilla Latent ODE reconstructions over-fit to the changepoint locations observed in training data. It is difficult for vanilla Latent ODEs to generalize on permutations of the piece-wise hybrid training trajectories, because they need to encode all sequence information into a single latent initial state. When a permutation in the sequence of SDFs is encountered, the non-robust latent representation predicts arbitrary dynamical shifts. The vanilla Latent ODE performs badly even in the zero change point reconstruction in Figure \ref{fig:lv_results} where one might expect it to do well, likely because it is anticipating potential change points. In contrast, the structured nature of the LatSegODE bypasses this need to learn a complex latent representation. Segmenting trajectories into SDFs allows for a complex hybrid trajectories to be represented by a sequence of simpler dynamics, yielding the accurate reconstructions shown.

\begin{figure}[ht]
    \centerline{\includegraphics[width=0.48\textwidth]{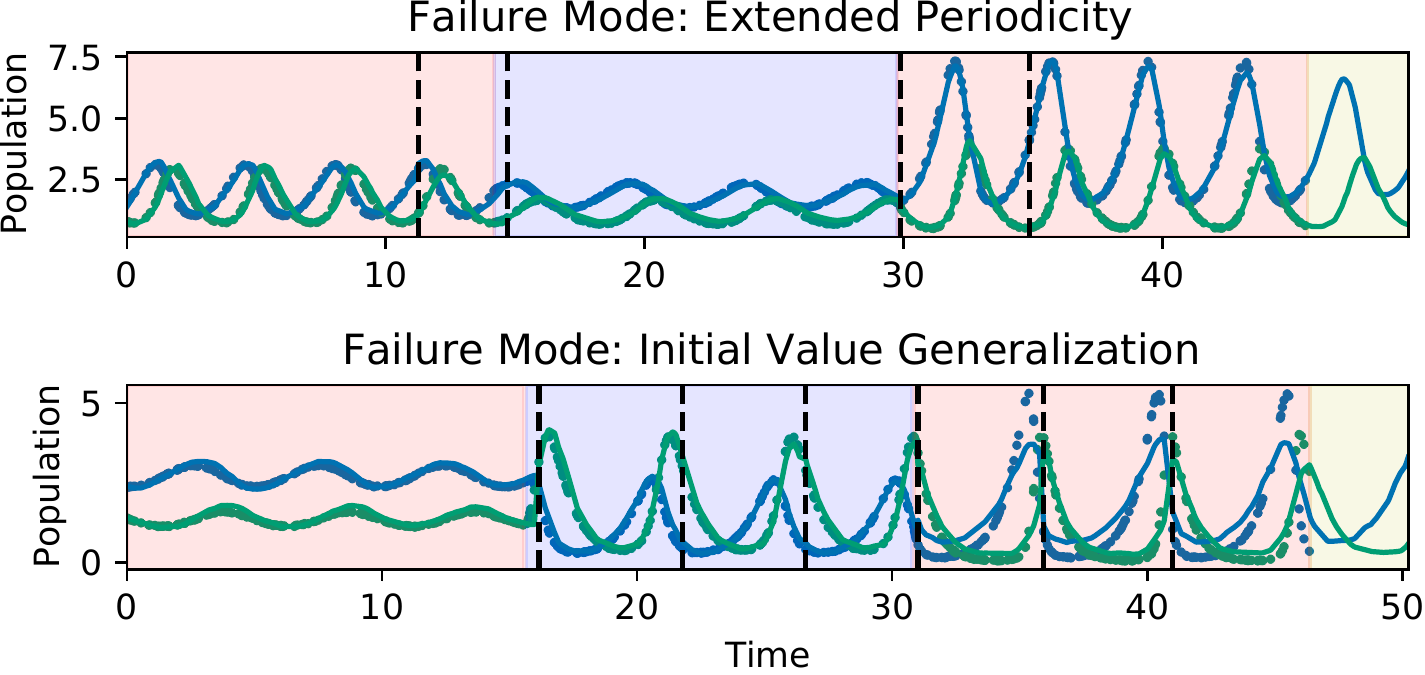}}
    \caption{Example failure modes encountered in Lotka Volterra modelling. See Figure \ref{fig:lv_results} for legend.}
    \label{fig:lv-fail}
\end{figure}

While the base model Latent ODE can powerfully represent SDFs, the LatSegODE method also inherits limitations of the architecture. We visualize two common failure modes in Figure \ref{fig:lv-fail}. We observed that learning limit cycles was challenging for the base model Latent ODE. In the top trajectory, imprecise base model representations cause deviation from the true periodic solution as time progresses. Eventually, enough error accumulates such that the accuracy gain from introducing a new segment overcomes the complexity cost of this action, resulting in over-segmentation. In the bottom trajectory, a failure mode is caused by the inability for the base model to generalize. Over-segmentation occurs if test trajectories contain SDFs which start outside of initial values founds in training data, such as at the second true changepoint. The base model cannot generalize well to unseen dynamical modes or initial values, so changepoints are erroneously introduced to improve fit. In Appendix I, we report data augmentation tricks which slightly improve generalization, remedying these issues.

This experiment also shows how the LatSegODE can be used in conjunction with physical simulators in a paradigm similar to simulator based inference \cite{cranmer2020frontier}. We train a MLP to map latent initial states from a trained base model to the labelled Lotka-Volterra coefficients of training SDFs. On test trajectories where the correct number of changepoints were predicted, we could recover the dynamical coefficients with a MSE of $\mathbf{0.08 \pm 0.01}$. In contexts such as Wright-Fisher population dynamics \cite{fisher1923xxi, wright1931evolution}, where forward simulation is available but cannot be expressed in closed form, the LatSegODE could be applied to solve inverse parameter estimation problems.

\subsection{UCI Character Trajectories}
Finally, we apply the LatSegODE to the UCI Character Trajectory data set \cite{Dua:2019}. This data set contains 2858 pen tip trajectories collected while writing letters of the alphabet. The trajectories are three dimensional, corresponding to x / y coordinates and pen pressure while writing one character. The data set is pre-processed by normalization and smoothing. Trajectories are regularly sampled with a maximum of 205 observations. We sanitized the data set by removing sections at the beginning and end of trajectories where no movement occurs. We use $5\%$ of the data for validation, and hold out $5\%$ for testing. The LatSegODE base model is trained on the remaining data, using each character trajectory as a SDF. Model architecture and hyper-parameters are reported in Appendix J.

We synthetically construct hybrid test trajectories by composing character trajectories. We randomly sampled a base character trajectory from the test set, then append up to two further randomly sampled character trajectories. To increase task difficulty, we add independent Gaussian noise with standard deviation of $0.2$. We also sub-sample the test trajectories to reduce number of observations the Latent ODE base model is able to condition upon. Using this method, we generate $75$ synthetic test hybrid trajectories, each containing zero to two changepoints. We report LatSegODE's segmentation performance on this synthetic test set in Table \ref{tab:ct-results}. 

\begin{table}[ht]
\caption{Segmentation results on UCI Character Trajectories.}
\label{tab:ct-results}
\begin{center}
\begin{small}
\begin{sc}
\begin{tabular}{lcccr}
\toprule
Method & \multicolumn{1}{p{1cm}}{\centering Rand\\Index} & \multicolumn{1}{p{2cm}}{\centering Hausdorff\\Metric} & \multicolumn{1}{p{1cm}}{\centering F1\\Score} \\
\midrule
LatSegODE & \textbf{0.9732} & \textbf{4.493} & \textbf{0.977} \\
\midrule
RPT-RBF & 0.7956 & 84.7 & 0.656\\
RPT-AR   & 0.6994 & 164.65 & 0.738\\
RPT-NORM   & 0.7693 & 105.92 & 0.611\\
\bottomrule
\end{tabular}
\end{sc}
\end{small}
\end{center}
\end{table}

In Figure \ref{fig:ct}, we provide an example reconstruction of a hybrid trajectory constructed by composing six character trajectories sampled from the test set. In both this figure and Table \ref{tab:ct-results}, the LatSegODE performs well in reconstructing long sequences of realistic data with noise, and accurately detects position of change in dynamical mode.

\section{Scope and Limitations}
\textbf{Data Labelling}: The LatSegODE requires SDF training data, typically obtained by splitting hybrid trajectories using labelled changepoints. This can be hard to obtain, so ideally, LatSegODE could be extended so it could be trained directly on hybrid trajectories. One approach would be marginalizing over changepoints during training using an inference procedure or a iterated-conditional-modes-like procedure that iterates between estimating an optimal segmentation given the current base model, and updating the base model given the segmentation.

\textbf{Dependency on Dynamical Models}: The LatSegODE relies on a Latent ODE base model to capture SDF behavior. Thus, it inherits many limitations of Latent ODEs, but any future advancements in the architecture and training of Latent ODEs can be directly integrated. While we chose to use Latent ODEs due its powerful representational ability, it could be replaced with any model for which marginal likelihood can be computed. Bayesian approaches to Neural ODEs such as the ODE2VAE \cite{yildiz2019ode} and Neural ODE Process \cite{norcliffe2021neural}, as well as the Latent SDE \cite{li2020scalable} method, could replace the Latent ODE base model with modifications. Thus, our framework can be used a paradigm for an expanded family of methods which combine PELT and dynamical models.

\textbf{Runtime}: The runtime of the LatSegODE can be improved. The current implementation naively computes the ODE solution for the union of batch timepoints. \citet{chen2020learning} provide a change of variables method to solve ODEs with irregular timepoints in parallel. This can reduce the memory bottleneck of the current approach, allowing additional parallelism to decrease evaluation runtime. The LatSegODE can integrated with recent methods to regularize ODE dynamics \cite{kelly2020learning}, \cite{finlay2020train}, which decrease evaluation runtime.

\section{Conclusion}
Here, we present the LatSegODE which leverages Latent ODEs to represent hybrid trajectories. Using a Latent ODE base model trained on SDFs and the PELT changepoint detection algorithm, we identify positions of jump discontinuity and switching dynamical mode, and restart latent dynamics from new initial states at these points. We provide a novel integration of Latent ODEs and CPD methods that uses the marginal likelihood of segments as a scoring function. We find that this Bayesian Occam's Razor effect prevents over-segmentation. We compared LatSegODE to baselines on synthetic and semi-synthetic benchmarks. Through qualitative analysis of example reconstructions, we highlight LatSegODE's ability to represent hybrid trajectories, and demonstrate common failure modes. The LatSegODE outperforms all baselines in both reconstruction and segmentation, supporting it as a novel approach to modelling hybrid trajectories governed by hybrid systems.

\section*{Acknowledgements}
We thank Tianxing Li and David Duvenaud for their helpful feedback and preliminary reviewing. We also thank Haoran Zhang, Yulia Rubanova, and other members of the Morris Lab for many helpful suggestions. Finally, we thank the ICML reviewers of this paper for their insightful feedback. Resources used in preparing this research were provided, in part, by the Memorial Sloan Kettering Cancer Center, Province of Ontario, the Government of Canada through CIFAR, and companies sponsoring the Vector Institute \url{www.vectorinstitute.ai/partners}.

\nocite{bengio2015scheduled}
\nocite{dupont2019augmented}
\nocite{fu2019cyclical}
\nocite{kidger2020hey}
\nocite{kingma2014adam}
\nocite{rackauckas2020universal}

\bibliography{main}
\bibliographystyle{style/icml2021}

\end{document}


\appendix

\section{PELT Algorithm}
The pruned exact linear time (PELT) algorithm is shown below:

\begin{algorithm2e}

\DontPrintSemicolon
\KwIn{Data observations $x_{1:N}$. Cost Function $\mathcal{C}$. Penalization $\beta$. Pruning parameter $K$.}
\KwOut{$cp$, position of changepoints.}

Initialize $F(0) = -\beta$. $cp = \{\}$, $R_1 = \{0\}$\;

\For{$\tau^* = 1, ..., N$}{
    Calculate $F(\tau^*) = \min_{\tau \in R_{\tau^*}} [F(\tau) + \mathcal{C}(x_{(\tau+1) : \tau^*}) + \beta]$ \;
    
    Let $\tau' = \arg \min_{\tau \in R_{\tau^*}}[F(\tau) + \mathcal{C}(x_{(\tau + 1):\tau^*} + \beta)$ \;
    
    Append $\tau'$ to $cp$ \;
    
    Set $R_{\tau^* + 1} = \{\tau \in R_{\tau^*}\cup \{\tau^*\} : F(\tau) + \mathcal{C}(x_{(\tau+1):\tau^*}) + \beta \leq F(\tau^*)$
}
\Return $cp$
\caption{PELT Algorithm}
\end{algorithm2e}

Array $R$ stores the changepoints under consideration for the optimal segmentation. Line 6 contains the pruning condition of PELT. After computation of the optimal sub-solution, PELT removes changepoints which cannot be optimal, as determined by equation (16-17) of the main text. In LatSegODE, we flip this algorithm to perform maximization instead of minimization by switching the relevant signs. We set $\beta$ to be zero, and use the marginal likelihood from equation (15) of the main text as $\mathcal{C}$. The choice of $K$ is problem dependent. Increasing K increases the threshold required to prune a changepoint. This increases segmentation accuracy at the cost of runtime.

\section{Runtime}
\begin{figure}[ht]
    \centering
    \begin{subfigure}[b]{0.45\textwidth}
        \includegraphics[width=\textwidth]{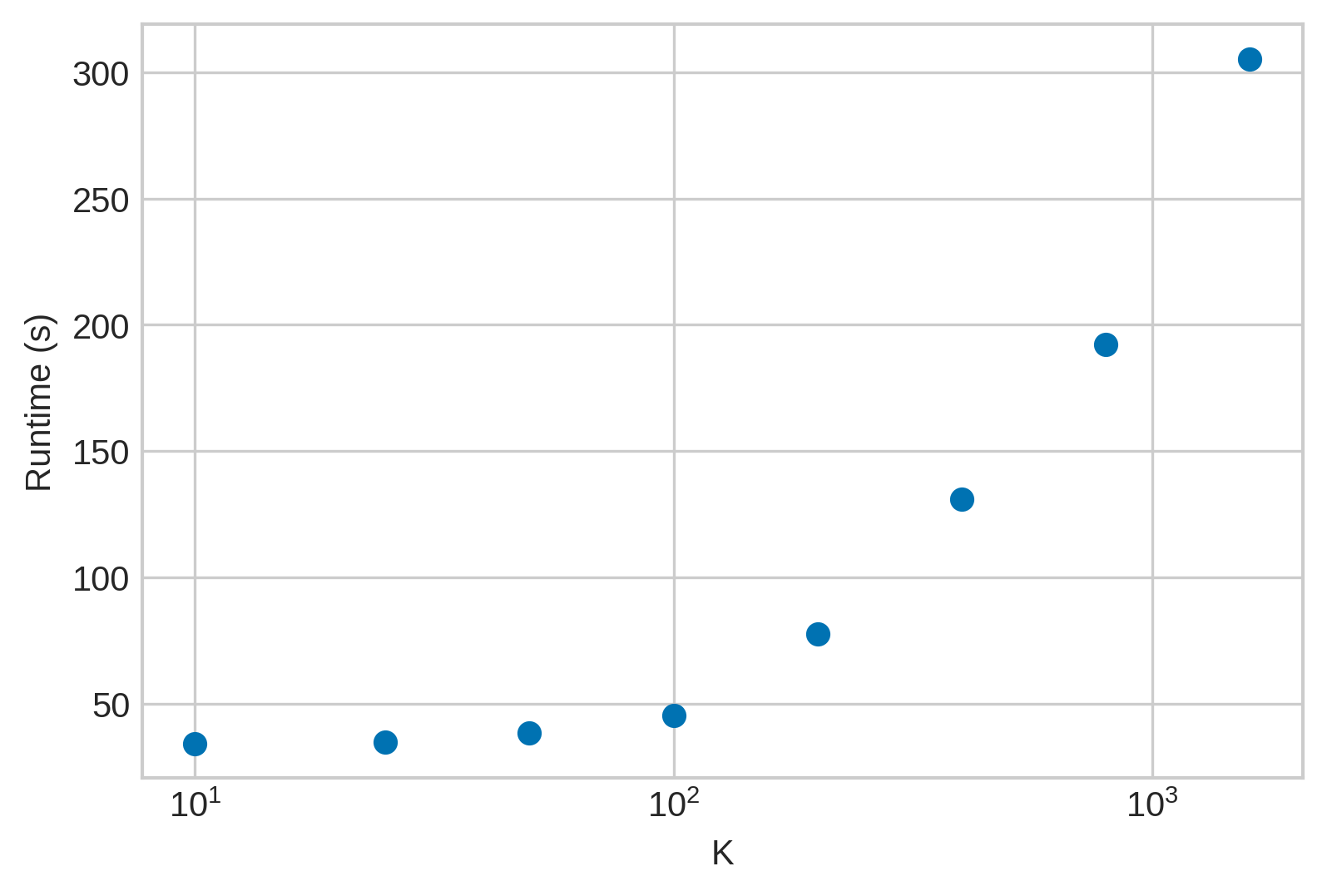}
        \caption{One ground truth changepoint.}
    \end{subfigure}
    \begin{subfigure}[b]{0.45\textwidth}
        \includegraphics[width=\textwidth]{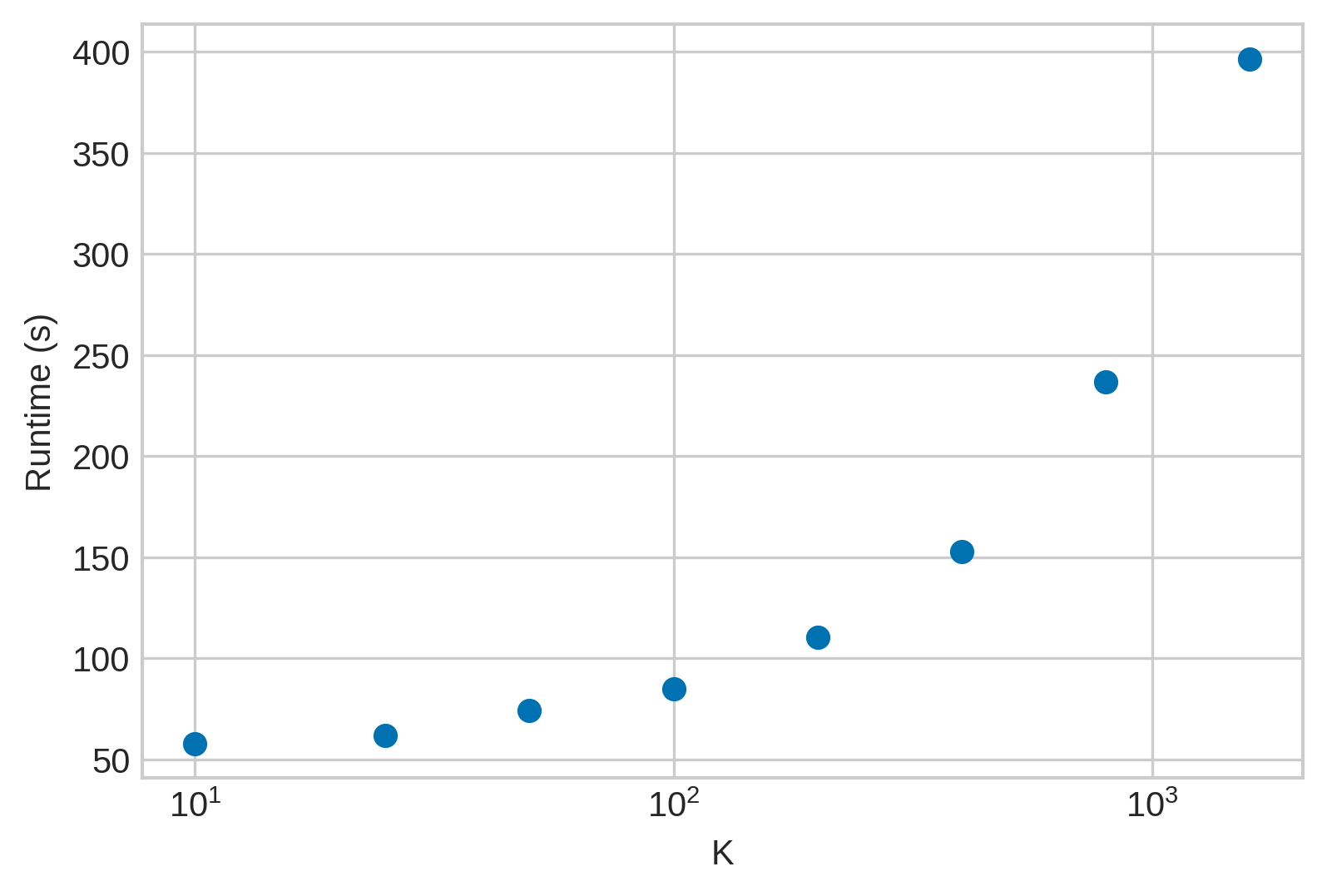}
        \caption{Two ground truth changepoints.}
    \end{subfigure}
    
    \caption{Run time as a function of $K$. X axis is in log scale.}
    \label{fig:rt}
\end{figure}

Here, we document the trade-off between $K$ and segmentation runtime. Setting a high $K$ allows more changepoints to be considered for the optimal segmentation. This means an optimal changepoint will be accidentally pruned with less frequency. Naturally, considering additional changepoints increases runtime, which we visualize in Figure \ref{fig:rt}.

We vary the $K$ term, and measure the wall runtime of LatSegODE segmentation. In subplot (a), we segment trajectories from the Lotka Volterra data set with one changepoint. These trajectories contains $410$ observations on average. In subplot (b), we segment trajectories from the Lotka Volterra data set containing two changepoints containing $620$ observations on average. We observe that segmentation accuracy increases as we increase $K$. This relationship is shown in Table \ref{apptab:Kseg}. Accuracy when using low values of $K$ is very poor. Accuracy rapidly increases as $K$ is increased, and then plateaus. For the Lotka-Volterra benchmark, a $K$ value around 50-100 offers the best performance for its runtime.

\begin{table}[ht]
\centering
\resizebox{0.75\textwidth}{!}{%
\begin{tabular}{l|llll|llll|}
\cline{2-9}
 & True CP = 1 &  &  &  & True CP = 2 &  &  &  \\ \hline
\multicolumn{1}{|l|}{K} & \begin{tabular}[c]{@{}l@{}}Rand \\ Index\end{tabular} & \begin{tabular}[c]{@{}l@{}}Hausdorff \\ Metric\end{tabular} & \begin{tabular}[c]{@{}l@{}}F1 \\ Score\end{tabular} & \begin{tabular}[c]{@{}l@{}}Annot.\\ Error\end{tabular} & \begin{tabular}[c]{@{}l@{}}Rand \\ Index\end{tabular} & \begin{tabular}[c]{@{}l@{}}Hausdorff \\ Metric\end{tabular} & \begin{tabular}[c]{@{}l@{}}F1 \\ Score\end{tabular} & \begin{tabular}[c]{@{}l@{}}Annot.\\ Error\end{tabular} \\ \hline
\multicolumn{1}{|l|}{10} & 0.7781 & 131.6 & 0.5578 & 1.4 & 0.8662 & 150.8 & 0.5714 & 0.133 \\
\multicolumn{1}{|l|}{25} & 0.8344 & 103.8 & 0.8133 & 0.4 & 0.9342 & 105.8 & 0.6929 & 1.0 \\
\multicolumn{1}{|l|}{50} & 0.9024 & 66.4 & 0.9333 & 0.2 & 0.9585 & 24.6 & 0.5524 & 0.6 \\
\multicolumn{1}{|l|}{100} & 0.9016 & 66.6 & 0.9333 & 0.2 & 0.9659 & 19.6 & 0.5619 & 0.4 \\
\multicolumn{1}{|l|}{200} & 0.9031 & 66.2 & 0.9333 & 0.2 & 0.9582 & 22.4 & 0.4952 & 0.4 \\ \hline
\end{tabular}%
}
\caption{Segmentation accuracy as a function of $K$. Metrics averaged over 5 runs.}
\label{apptab:Kseg}
\end{table}

\section{Neural Event ODEs}
\begin{figure}[ht]
    \centering
    \includegraphics[width=0.5\textwidth]{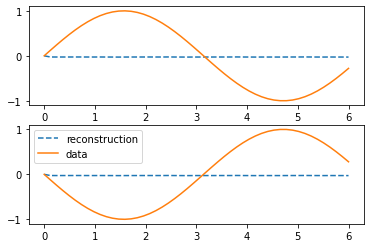}
    \caption{An example of trajectories which cannot be simultaneously represented by a single Neural Event ODE. Both trajectories begin with an initial value of zero, but evolve with opposite dynamics.}
    \label{fig:eventode}
\end{figure}

Here, we demonstrate our intuition on why the Neural ODE and Neural Event ODE methods are unable to converge within our problem domains. Neural ODEs represent trajectories as a deterministic function of its initial state. Furthermore, two different trajectories cannot evolve from an identical initial state. In experimental benchmarks, and time series modelling in general, trajectories may start at identical initial states but later diverge. Neural ODEs and Neural Event ODEs, which can only encode one trajectory per initial state, are not designed to represent this class of time series. Consequently, we observe that Neural ODEs and Neural Event ODEs do not converge during training on our benchmarks. An example is shown in Figure \ref{fig:eventode}.

The figure shows two sine waves with opposite amplitudes a shared initial value of zero. As expected, we see that Neural Event ODE reconstructions are poor, and did not converge during training. The Neural Event ODE was trained using the sine wave benchmark data set. The ODE function is parameterized using a 3 layer neural network (NN) with Tanh activations and 256 units in hidden layers. We use a 2 layer NN with 256 width and ReLU activations to parameterize the update and event functions. Hyper-parameter search was performed by varying the number of layers and hidden layer width in the ranges (2, 3) and (64, 128, 256). 

Latent ODEs does not suffer this representational limitation. We consider the example in the figure. The Latent ODE can represent this example by using two different latent trajectories which begin at non-identical latent initial states. The decoder neural network can then learn a surjective function which maps the non-identical latent initial states to identical initial values in data space. Extension of the Neural Event ODE to a latent architecture is a promising direction to circumvent limitations.

\section{General Experimental Design}
All experiments are run on a single Titan Xp GPU. Due to computational constraints, each training run was only performed once per hyper-parameter choice. Each segmentation evaluation was also only run once due to computational constraints.

\subsection{Reconstructive Baselines}
Reconstructive performance was baselined using several methods. First, we selected the GRU model, a standard choice in representing time series. The GRU $\Delta t$ model uses the time delta between observations as an additional input feature. Inclusion of the time delta improved extrapolation performance. The GRU-ODE model uses a Neural ODE to evolve hidden state between GRU updates. All of these methods are trained in an auto-regressive manner. During training, scheduled sampling \cite{bengio2015scheduled} is used. At each time step, with $25\%$ probability, the input data is replaced with the prediction output of the previous time point. We also mask $25\%$ of the observations at the end of the trajectory (in addition to masking described in main text), and include the prediction loss on these points. These models are trained using the mean squared error loss between ground truth and predicted data points.

As the LatSegODE is provided changepoint positions (and thus isolated simple dynamical flows) during training, we also provide this information to baseline methods. We trained models on trajectories which have been split into simple dynamical flows (SDFs), and also separately trained on entire hybrid trajectories. The best performing model resulting from these two training configurations is reported.

We trained vanilla Latent ODEs using only whole hybrid trajectories. Obtaining convergence was difficult, and required days of training. We use the training strategy from \cite{rackauckas2020universal}, where we iterative grow the learned reconstruction. We experimented using an augmented Neural ODE \cite{dupont2019augmented} to represent latent dynamics, but did not observe significant benefit. We applied an adjoint computation modification to speed up training \cite{kidger2020hey}, but did not perform rigorous empirical testing to evaluate its benefits.

We hacked data generation to enable faster experimentation. Generated observation times were irregularly sampled, but aligned for all trajectories. This enables batch training. In practice, Latent ODEs can handle non-aligned irregularly sampled time series by solving for the union of all time points in a mini-batch. This is drastically increases memory usage and runtime. Thus, for convenience, we adopt the aforementioned hack.

\subsection{Segmentation} 
We report the specific formulations of the CPD methods and metrics used in our benchmarks. These descriptions are referenced from the \verb+ruptures+ \cite{truong2020selective} package documentation.

\subsubsection*{Metrics}
The Rand Index \cite{rand1971objective} measures the overlap between the predicted segmentation and the ground truth segmentation for a segmentation $S$ on data points $x_{1:T}$. A membership matrix $A$ is defined such that $A_{ij} = 1$ if $x_i$ and $x_j$ are in the same segment. Otherwise, $A_{ij} = 0$. Membership matrices are generated for both the ground truth segmentation $(A)$ and the predicted segmentation $(\Tilde{A})$. The Rand index is defined as:

\begin{equation}
    \text{RandIndex} = \frac{\sum_{i<j} \mathds{1}[A == \Tilde{A}]}{T(T-1) / 2}
\end{equation}

The Hausdorff metric \cite{rockafellar2009variational} is a measure of the maximal distance between the predicted segmentation and the ground truth segmentation. Given a set of ground truth changepoints $t_1, t_2, ...$ and predicted changepoints $\hat{t_1}, \hat{t_2}, ...$, it is computed as:
\begin{equation}
    \text{Hausdorff}(\{t_k\}_k, \{\hat{t}_k\}_k) = \max\{ \max_k \min_l |t_k - \hat{t}_l|, \max_k \min_l | t_l - \hat{t}_k |\}
\end{equation}
Intuitively, it returns the max of the set of distances from each predicted changepoint to their closest ground truth changepoint.

The F1 score is calculated as the standard F1 score, namely the harmonic mean between precision and recall:
\begin{equation}
    F_1 = 2 \times \frac{\text{precision} \times \text{recall}}{\text{precision} + \text{recall}}
\end{equation}
A changepoint prediction is considered correct if a falls within 10 indices of a true changepoint. This definition of correctness is used to calculate precision and recall for the F1 score.

The annotation error reports the difference between the count of predicted changepoints, and the count of true changepoints.

\subsubsection*{Methods}
We outline the three cost functions used in the main text. Other cost functions (L1/L2 deviation, rank transformation, Mahalanobis distance) were evaluated not reported due to inferior performance. These cost functions are used with PELT. Alternative search methods such as binary segmentation and sliding windows were not considered. They are prone to returning sub-optimal results, as they are greedy methods.

\textbf{RPT-RBF}: This cost function detects changes in the distribution of a sequence of i.i.d. random variables. It introduces a kernel $k(., .): \mathbb{R}^d \times \mathbb{R}^d \rightarrow \mathbb{R}$ and a feature map $\Phi: \mathbb{R}^d \rightarrow \mathcal{H}$ where $\mathcal{H}$ is a Hilbert space. RPT-RBF embeds a signal as $\{\Phi(x_t)\}_t$, and detects changes in the mean so that the cost function on an interval $I$ is:
\begin{equation}
    c(x_I) = \sum_{t\in I} ||\Phi(x_t) - \Bar{\mu}||^2_\mathcal{H}
\end{equation}
where $\Bar{\mu}$ is the empirical mean of the sub-trajectory $\{\Phi(x_t)\}_{t\in I}$. RPT-RBF uses the radial basis function, which is defined as:
\begin{equation}
k(x, y) = \exp(-\gamma ||x - y||^2)
\end{equation}
where $||.||$ is the Euclidean norm and $\gamma > 0$ is a smoothing parameter known as the bandwidth. Two other kernels are provided, which use a cosine similarity kernel, and a linear model. We use the RBF as it provided better performance.

\textbf{RPT-NORM}: This cost function scores a segment using a sequence of multivariate Gaussian random variables, i.e., a Gaussian Process. For a signal $\{x_t\}_t$ on interval $I$, this cost is defined as:
\begin{equation}
    c(x_I) = |I| \log \det \hat{\Sigma}_I
\end{equation}
where $\hat{\Sigma}_I$ is the empirical covariance matrix of the data points $\{x_t\}_{t\in I}$.

\textbf{RPT-AR}: This cost function introduces an auto-regressive model. It represent unknown changepoint indices as $0 < t_1 < .. < t_N$. A piece-wise auto-regressive model is introduced:
\begin{equation}
    x_t = z'_t \delta_j + \epsilon_t \hspace{2em} \forall t = t_j, ..., t_{j+1} - 1
\end{equation}
To clarify, $t_j$ is the segmentation boundary, while $t$ represents actual time indices. Thus, $j > 1$ represents the number of a segment. The variable $z_t = [x_{t-1}, ... x_{t-p}]$ is the lag vector, with $p$ being the order of the auto-regressive model.

On an interval $I$, the cost is defined as:
\begin{equation}
    c(x_I) = \min_{\delta \in \mathbb{R}^p} \sum_{t\in I} || x_t - \delta' z_t ||^2_2
\end{equation}

For this cost function, the lag term is a hyper-parameter. We chose the default value (ten) and a small grid search around this value confirmed it performs the best.

\section{Sine Wave Experimental Setup}
We simulate 7500 total trajectories, of which 300 is used for validation and 150 for test. Each trajectory can contain up to two changepoints. The time length and number of observations in each SDF between changepoints is uniformly sampled from the ranges $(3,5)$ and $(50, 150)$ respectively. The amplitude and frequency of SDFs are uniformly sampled from the ranges $(-8, 8)$ and $(2,4)$, respectively. The phase of each SDF is also randomly drawn, which introduces jump discontinuities. The time points of observation were randomly sampled using a uniform distribution on the range of the segment, but were forcibly aligned for experimental convenience. The absolute change in amplitude between SDFs is at least $2.5$, and we added independent Gaussian noise with standard deviation $0.025$ to simulate noise. At evaluation time, all segmentation methods use a minimum segment length of $20$, and we use a $K$ term of $200$.

Next, we report architecture hyperparameters and training procedure. Unless otherwise stated,  hyper-parameter search was mainly performed on by adjusting the depth and width of NNs used to parameterize the Neural ODEs in various architectures. We start with a 3 layer NN with 50 units, and increase width/depth until the model can fit the data.

\textbf{GRU / GRU$\Delta$t} The GRU model used 100 units in the GRU, and a 20 dimensional hidden state. A 2 layer NN with 100 units and ReLU activations is used for the output network. The GRU $\Delta$t model used an identical architecture. These hyperparameters were selected using grid search. We started with 50 units in the GRU, and increased number of units until overfit occurred. Both models were trained using Adamax \cite{kingma2014adam}, with a learning rate of 0.01. Learning rate was manually decayed to 1e-3 and then 1e-4 when loss plateaued. A batch size of $512$ was used. This was selected based on memory constraints, as larger batch sizes allowed for faster convergence.

\textbf{GRU-ODE} The GRU-ODE model used 100 units in the GRU, and a 10 dimensional hidden state. The output network is a 2 layer 100 unit NN with ReLU activations. The Neural ODE component of the GRU-ODE is parameterized by a 3 layer 100 unit NN with Tanh activations. We found that using batch size 1 provided good results, at the cost of a very long training run. The model was optimized using Adamax, with manual learning rate reduction on plateaus from 0.01 to 1e-3 to 1e-4.

\textbf{Latent ODE} The vanilla Latent ODE model used a encoder with a 32 dimensional hidden state. The latent dynamics were 16 dimensional. The dimensionality in the vanilla Latent ODE is larger than the base model used in the LatSegODE, as the latent initial state for vanilla Latent ODEs must encode more information. The encoder GRU contains 100 units, and the encoder Neural ODE was parameterized by a 3 layer 100 units NN with Tanh activations. The Neural ODE representing latent dynamics was parameterized identically. We used a 2 layer 100 unit NN with ReLU activations as the decoder. The Latent ODE is trained using Adamax with learning rate 0.01. The learning rate was manually decayed to 1e-3 when loss plateaued. A batch size of 128 was used. We used KL annealing \cite{fu2019cyclical}, such that the KL weight was 0 at the start of training, and reached 1 at epoch 5. The fixed variance used to compute the ELBO was set to 1. Latent dynamics were solved using the dopri5 solver using relative and absolute tolerances of 1e-8 and 1e-8. The encoder Neural ODE was solved using Euler's method.

\textbf{LatSegODE} The LatSegODE used a 10 dimensional hidden state in the encoder, and a 5 dimensional latent dynamical state. The original Latent ODE paper observed that the dimension of the encoder hidden state must be larger than the dimension of the latent dynamical state. Our experimentation supported this claim, and training was not possible if this rule was violated. The GRU in the GRU-ODE encoder uses 100 units, and the encoder Neural ODE is a 2 layer 100 unit NN with Tanh activations. The Neural ODE used to represent latent dynamics used the same hyperparameters. We found that Tanh activations were required for stable training in Latent ODEs. We attempted other activations such as Swish, ReLU, and Softplus, but found this caused numerical under/overflow. We used a 2 layer 100 unit decoder network with ReLU activations.

It is important to use ReLU activations in the decoder network. Alternative activations such as the Tanh or SoftPlus are bijective activations, meaning the initial observed data point in data space can never map to multiple latent initial states. Using the Tanh activation greatly limits the representational power of the Latent ODE base model when representing data trajectories which start at the same value, but later diverge.

The LatSegODE was trained using Adamax using learning rate 0.01, reduced to 1e-3 manually when validation loss plateaued. We used a batch size of 256. We used the dopri5 solver to solve latent dynamics with relative tolerance of 1e-5 and absolute tolerance of 1e-6. Increasing the ODE solve tolerance had no adverse effects on results and sped up training. The encoder Neural ODE used Euler's method to solve dynamics. We used KL annealing during training. 

The fixed variance used to compute ELBO is set to 1. At segmentation time, it was critical to use the same fixed variance to evaluate marginal likelihood. We used $100$ Monte Carlo samples to estimate marginal likelihood. We used a $K$ term of $200$. To reduce the memory cost of segmentation, we rounded times of observation to the nearest 0.01, reducing the size of the union of time points which must be solved.

\pagebreak

\section{Sine Wave Results}
Experimental results on the Sine Wave dataset. The Latent ODE trained on input augmented with a binary time series of changepoint locations is denoted Aug. Latent ODE.
\begin{table}[ht]
\centering
\resizebox{0.75\textwidth}{!}{%
\begin{tabular}{l|lll|llll|}
\cline{2-8}
 & \multicolumn{3}{l|}{Reconstruction} & \multicolumn{4}{l|}{Segmentation} \\ \hline
\multicolumn{1}{|l|}{Method} & \begin{tabular}[c]{@{}l@{}}Total \\ MSE\end{tabular} & \begin{tabular}[c]{@{}l@{}}Interp. \\ MSE\end{tabular} & \begin{tabular}[c]{@{}l@{}}Extrap. \\ MSE\end{tabular} & \begin{tabular}[c]{@{}l@{}}Rand \\ Index\end{tabular} ($\uparrow$) & \begin{tabular}[c]{@{}l@{}}F1 \\ Score\end{tabular} ($\uparrow$) & \begin{tabular}[c]{@{}l@{}}Hausdorff \\ Metric\end{tabular} ($\downarrow$) & \begin{tabular}[c]{@{}l@{}}Annot. \\ Error\end{tabular} \\ \hline
\multicolumn{1}{|l|}{LatSegODE} & \textbf{1.933} & \textbf{0.639} & \textbf{4.266} & \textbf{0.993} & \textbf{0.989} & \textbf{2.107} & 0.033 \\ \hline
\multicolumn{1}{|l|}{GRU} & 8.065  & 0.862 & 20.634 & - & - & - & - \\
\multicolumn{1}{|l|}{GRU $\Delta$t} & 6.911  & 0.973 & 17.560 & - & - & - & - \\
\multicolumn{1}{|l|}{GRU-ODE} & 6.616 & 0.815 & 17.130 & - & - & - & - \\
\multicolumn{1}{|l|}{Latent ODE} & 12.877 & 9.111 & 17.721 & - & - & - & - \\
\multicolumn{1}{|l|}{Aug. Latent ODE} & 13.273 & 4.914 & 26.922 & - & - & - & - \\ \hline
\multicolumn{1}{|l|}{RPT-RBF} & - & - & - & 0.918 & 0.837 & 25.220 & - \\
\multicolumn{1}{|l|}{RPT-AR} & - & - & - & 0.938 & 0.887 & 26.013 & - \\
\multicolumn{1}{|l|}{RPT-NORM} & - & - & - & 0.855 & 0.761 & 47.213 & - \\ \hline
\end{tabular}%
}
\caption{Evaluation on Sine Wave data set. Values are averages over 150 test trajectories. Arrows beside metric denote whether higher values ($\uparrow$) or lower values ($\downarrow$) indicate better performance.}
\label{tab:sine}
\end{table}
\section{Lotka-Volterra Experimental Setup}
Here, we report the parameters used to generate the Lotka Volterra data set, and the model hyper-parameters used during bench marking. As previously reported, we simulated $34000$ training hybrid trajectories, with $600$ trajectories used as a validation set, and $150$ used for test. Each trajectory contained zero to two changepoints. Changepoints in these trajectories are labelled, and the LatSegODE base model is trained on the SDFs between changepoints. The vanilla Latent ODE baseline is trained only on the hybrid trajectories, while other baselines were trained both on full hybrid trajectories and SDFs separately, with the best result reported.

Each SDF between changepoints was generated with between $175$ to $225$ observations, and ends at a time randomly sampled between $(14, 16)$. The coefficients for the Lotka-Volterra SDFs were uniformly sampled from ranges $(0.5, 1.5), (0.5, 1.5), (1.5, 2.5), (0.5, 1.5)$ for coefficients $\alpha, \beta, \delta, \gamma$ respectively. We enforced a minimum change in the norm of coefficients of $0.6$ between SDFs, and added independent Gaussian noise of $0.01$ to all trajectories. We masked $20\%$ of data points for interpolation testing, and the last $100$ data points for extrapolation testing, as reported in the main text.

In the next appendix, we separately evaluate performance on data sets containing jump discontinuity (JD) and without (SD). For the JD set, each SDF within a trajectory is restarted at a new initial population at changepoints, uniformly sampled from ranges $(1.5, 2.5), (0.5. 1.5)$ for $x, y$ respectively. In the SD set, SDFs are initialized from the same distribution, but remain continuous at subsequent changepoints. To clarify, trajectories in the SD set do not contain jump discontinuities at changepoints, and these locations only feature a switch in dynamical mode.

Next, we report the hyperparameters and training procedures in the Lotka Volterra experiments. We performed hyperparameter search over the width and depth of the NNs parameterizing Neural ODEs in our architectures. We started with a one layer 64 unit NN, and increased the number of layers and units until convergence was possible. Latent dimension selection was difficult. We started with a latent dimension of 64, and slowly decreased capacity until under-fitting occurred.

\textbf{GRU / GRU$\Delta$t} We found that the GRU always performed worse than the GRU$\Delta$t, so we did not report its results. The GRU $\Delta$t model used 16 dimensions in its hidden state, and 200 units in the GRU. A 2 layer 100 unit NN with ReLU activations was used as an output network. The model was trained using Adamax, where learning rate was decayed from 1e-2 to 1e-3 to 1e-4 each time validation loss plateaued. We used a batch size of 512. 

\textbf{GRU-ODE} The GRU-ODE model used a hidden state with dimension 50, and GRU with 100 units. A 2 layer 100 unit output network with ReLU activations was used. The Neural ODE was parameterized with using a 3 layer 200 unit NN with Tanh activations. The GRU-ODE was trained by Adamax using a constant 1e-3 learning rate. The Neural ODE solutions were solved using dopri5 with relative and absolute tolerances of 1e-3 and 1e-4.

\textbf{Latent ODE} Latent ODE models were very hard to train, and many approaches were taken. First, we attempted to tune hyperparameters, increasing both latent dimension size, and the number of parameters in NNs parameterizing Neural ODEs. The rationale was that the added complexity in hybrid trajectories necessitated increased model complexity. We found that model reconstruction accuracy in both data sets were poor even after many epochs. A more successful strategy was to adopt an iterative growing scheme for training \cite{rackauckas2020universal}. We also adopted a forward prediction training strategy. We started training on a masked sub-trajectory with only the first 50 data points, and evaluate loss on prediction of the next 50 data points. The training loss was computed using all observed points. After each epoch, we increased the number of observed data points by 50. We also found that taking multiple samples of the latent initial state during training could stabilize gradient estimates, and we used 3 samples per trajectory.

Latent ODEs used were trained using Adamax with a constant learning rate of 5e-3, decayed to 1e-3 after 10 epochs. A batch size of 256 is used. Latent ODEs were trained using KL annealing such that a KL weight of 1 was reached after 10 epochs. A fixed variance of 0.01 was used in the ELBO.

The Latent ODE models had a hidden state of dimension 16, and a latent state of dimension 8. The encoder used 100 GRU units, with its Neural ODE parameterized by a 3 layer 100 unit NN with Tanh activations. The identical hyper parameters are used for the latent Neural ODE neural network. We used a linear decoder. Latent dynamics were solved using dopri5 with relative and absolute tolerance of 1e-4 and 1e-4.

\textbf{LatSegODE} The LatSegODE used a base model with a hidden state of dimension 16, and a latent state with dimension 8. The encoder used GRUs with 100 units, and both the encoder Neural ODE and latent Neural ODE was parameterized by 3 layer 100 unit NNs with Tanh activations. A decoder network with 2 layer 100 units with ReLU activations was used. 

We trained our model using Adamax, with an initial learning of 5e-3. Learning rate was decayed to 1e-3 after 10 epochs, and reduced to 1e-4 after validation loss plateaued. A batch size of 256 was used. Latent dynamics were solved using dopri5, with relative and absolute tolerances of 1e-4 and 1e-4. Training used KL annealing such that KL weight started from 0 and reached 1 after 10 epochs. A fixed variance of 0.01 was used to compute the ELBO.

At segmentation time, we used a fixed variance of 0.01 to compute the marginal likelihood. We took 100 MC samples to compute the marginal likelihood, and rounded times of observation to 2 decimal places.

\pagebreak

\section{Lotka-Volterra Expanded Experiments}
We report the full results of Lotka-Volterra experiments on additional data sets, jump discontinuous (JD) and switching dynamics (SD). The JD set is identical to the set reported in the main text. The SD does not contain jump discontinuity at changepoints, and the trajectory only switches to a new dynamical mode. The SD set is included evaluate performance of the LatSegODE without discontinuous jumps, which may be theoretically easier to solve since ODE solves do not need to bridge a jump discontinuity. We find the opposite actually occurs, as the less distinct changepoints in the SD set yields a harder data set, shown by the decreased performance of baselines. We report the metrics in Table \ref{tab:lv-metrics} below.

\begin{table}[ht]
\centering

\begin{subtable}{\textwidth}
\centering
\resizebox{0.75\textwidth}{!}{%
\begin{tabular}{l|lll|llll|}
\cline{2-8}
 & \multicolumn{3}{l|}{Reconstruction} & \multicolumn{4}{l|}{Segmentation} \\ \hline
\multicolumn{1}{|l|}{Method} & \begin{tabular}[c]{@{}l@{}}Total \\ MSE\end{tabular} & \begin{tabular}[c]{@{}l@{}}Interp. \\ MSE\end{tabular} & \begin{tabular}[c]{@{}l@{}}Extrap. \\ MSE\end{tabular} & \begin{tabular}[c]{@{}l@{}}Rand \\ Index\end{tabular} ($\uparrow$) & \begin{tabular}[c]{@{}l@{}}F1 \\ Score\end{tabular} ($\uparrow$) & \begin{tabular}[c]{@{}l@{}}Hausdorff \\ Metric\end{tabular} ($\downarrow$) & \begin{tabular}[c]{@{}l@{}}Annot. \\ Error\end{tabular} \\ \hline
\multicolumn{1}{|l|}{LatSegODE} & \textbf{0.068} & 0.0312 & \textbf{0.2396} & \textbf{0.9464} & \textbf{0.8268} & \textbf{47.67} & 0.76 \\ \hline
\multicolumn{1}{|l|}{GRU $\Delta$t} & 0.1718 & \textbf{0.0193} & 0.8329 & - & - & - & - \\
\multicolumn{1}{|l|}{GRU-ODE} & 0.2747 & 0.1201 & 2.0358 & - & - & - & - \\
\multicolumn{1}{|l|}{Latent ODE} & 0.6155 & 0.5072 & 0.9505 & - & - &  & - \\ \hline
\multicolumn{1}{|l|}{RPT-RBF} & - & - & - & 0.7956 & 0.4167 & 84.7 & - \\
\multicolumn{1}{|l|}{RPT-AR} & - & - & - & 0.6994 & 0.4367 & 164.65 & - \\
\multicolumn{1}{|l|}{RPT-GPNorm} & - & - & - & 0.7693 & 0.42 & 105.92 & - \\ \hline
\end{tabular}%
}
\caption{Jump discontinuous (JD) test set.}
\end{subtable}
\vspace{1em}

\begin{subtable}{\textwidth}
\centering
\resizebox{0.75\textwidth}{!}{%
\begin{tabular}{l|lll|llll|}
\cline{2-8}
 & \multicolumn{3}{l|}{Reconstruction} & \multicolumn{4}{l|}{Segmentation} \\ \hline
\multicolumn{1}{|l|}{Method} & \begin{tabular}[c]{@{}l@{}}Total \\ MSE\end{tabular} & \begin{tabular}[c]{@{}l@{}}Interp. \\ MSE\end{tabular} & \begin{tabular}[c]{@{}l@{}}Extrap. \\ MSE\end{tabular} & \begin{tabular}[c]{@{}l@{}}Rand \\ Index\end{tabular} ($\uparrow$) & \begin{tabular}[c]{@{}l@{}}F1 \\ Score\end{tabular} ($\uparrow$) & \begin{tabular}[c]{@{}l@{}}Hausdorff \\ Metric\end{tabular} ($\downarrow$) & \begin{tabular}[c]{@{}l@{}}Annot. \\ Error\end{tabular} \\ \hline
\multicolumn{1}{|l|}{LatSegODE} & \textbf{0.2284} & 0.2176 & \textbf{1.010} & \textbf{0.8758} & \textbf{0.6857} & \textbf{75.66} & 2.42 \\ \hline
\multicolumn{1}{|l|}{GRU $\Delta$t} & 0.4519 & \textbf{0.0383} & 2.427 & - & - & - & - \\
\multicolumn{1}{|l|}{GRU-ODE} & 0.4386 & 0.2176 & 3.280 & - & - & - & - \\
\multicolumn{1}{|l|}{Latent ODE} & 1.4027 & 1.1924 & 2.1314 & - & - & - & - \\ \hline
\multicolumn{1}{|l|}{RPT-RBF} & - & - & - & 0.7833 & 0.4233 & 77.77 & - \\
\multicolumn{1}{|l|}{RPT-AR} & - & - & - & 0.7020 & 0.4433 & 138.24 & - \\
\multicolumn{1}{|l|}{RPT-GPNorm} & - & - & - & 0.7672 & 0.4233 & 94.44 & - \\ \hline
\end{tabular}%
}
\caption{Switching dynamical mode (SD) test set.}
\end{subtable}

\caption{Expanded results on Lotka Volterra hybrid trajectory benchmark. Values are averages over 150 test trajectories. Arrows beside metric denote whether higher values ($\uparrow$) or lower values ($\downarrow$) indicate better performance.}
\label{tab:lv-metrics}
\end{table}

\section{Data Augmentation Strategies}
We report two data augmentation techniques which can increase training speed and the ability for the Latent ODE to generalize. These techniques are applied on data batches prior to each training iteration. First, we propose sub-sampling trajectories by randomly removing a percentage of observed points. This method resembles techniques which iteratively grows trajectory length throughout training to avoid local minima \cite{rackauckas2020universal}, and we hypothesize sub-sampling confers a similar benefit. We also introduce start-truncation, where the first $N$ data observations are cropped. We hypothesize this augmentation decreases generalization error by exposing the Latent ODE encoder to more initial states.

\begin{figure}[ht]
    \centering
    \begin{subfigure}{0.4\textwidth}
        \includegraphics[width=\textwidth]{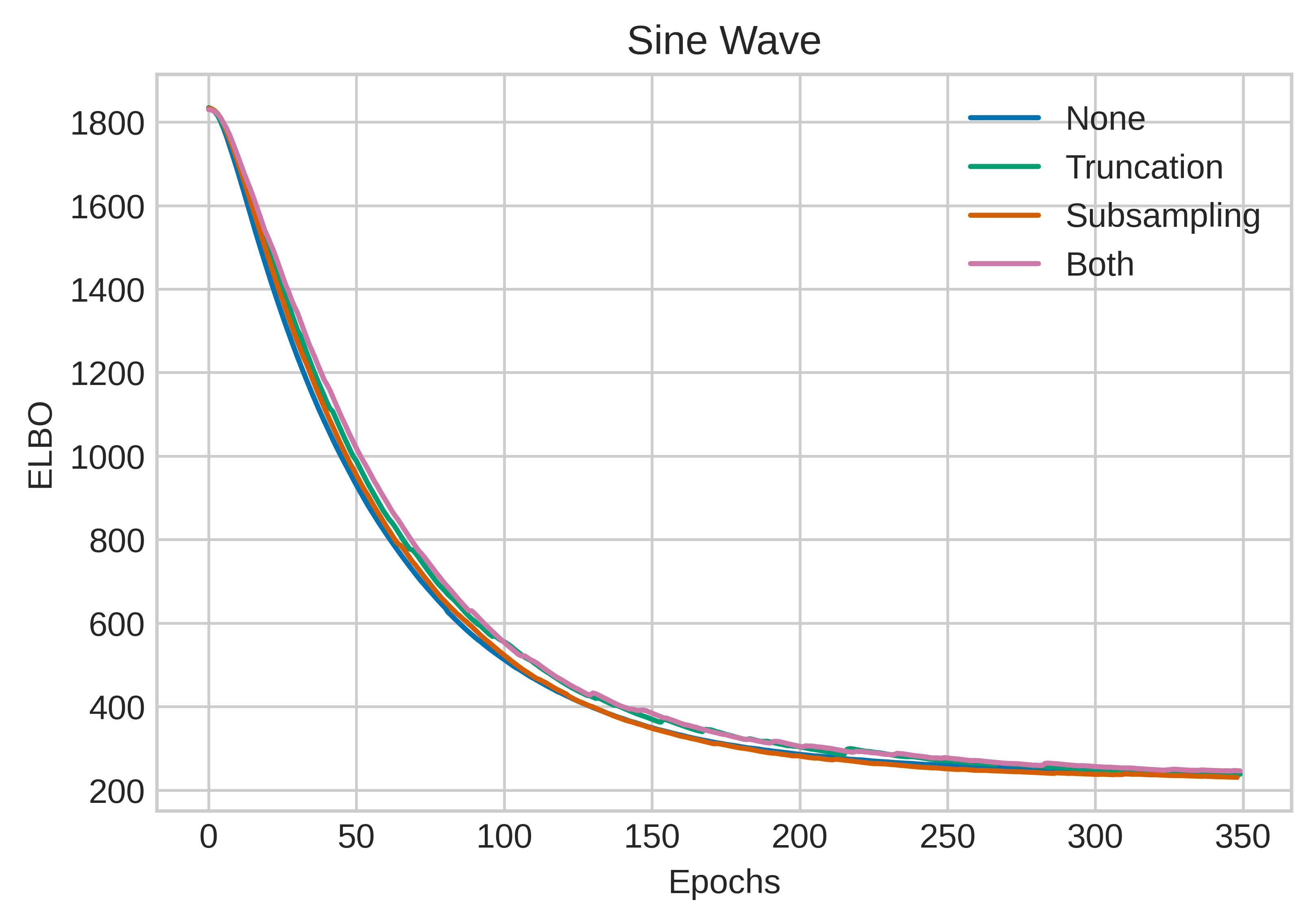}
        \caption{}
    \end{subfigure}
    \begin{subfigure}{0.4\textwidth}
        \includegraphics[width=\textwidth]{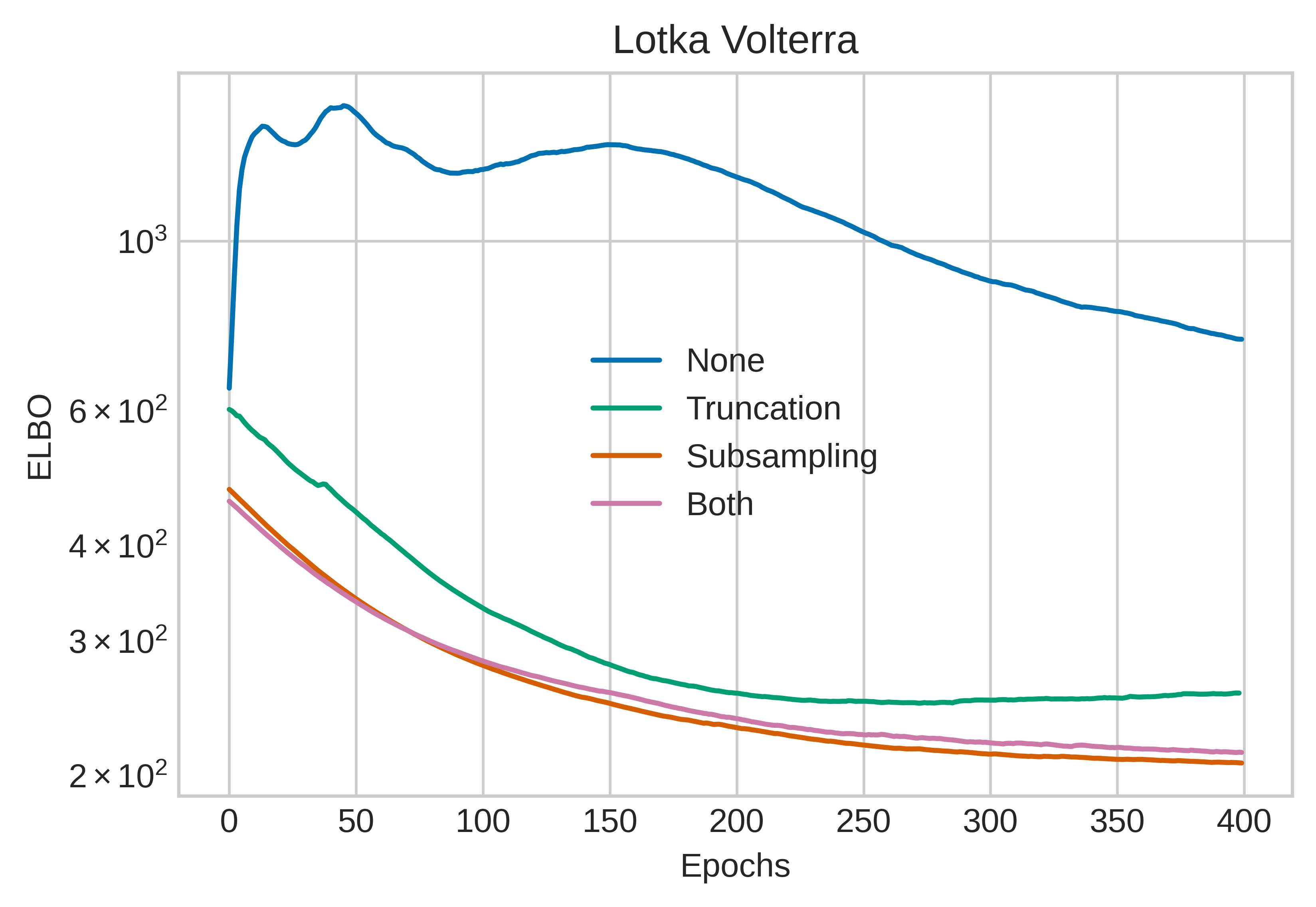}
        \caption{}
    \end{subfigure}
    \caption{Validation loss curves using data augmentations. Plot (a) shows runs on Sine Wave data, while (b) shows runs on Lotka Volterra data.}
    \label{fig:loss_curves}
\end{figure}

To demonstrate the efficacy of these augmentation methods, we generate $10000$ SDFs from each of the previous experimental domains, with $500$ trajectories held for validation and $500$ for test evaluation. Each trajectory contains $200$ samples. For sub-sampling, we randomly select between $40$ to $200$ data points to train per batch. For truncation, we randomly select the number of data points to remove from range $(0, 160)$. We train Latent ODE models with and without augmentation techniques, and plot the validation loss curve for each training run in Figure \ref{fig:loss_curves}. 

\begin{table}[ht]
\centering
\resizebox{0.47\textwidth}{!}{%
\begin{tabular}{l|ll|ll|}
\cline{2-5}
 & \multicolumn{2}{l|}{Sine Wave} & \multicolumn{2}{l|}{Lotka Volterra} \\ \hline
\multicolumn{1}{|l|}{Method} & Test MSE & \begin{tabular}[c]{@{}l@{}}Runtime \\ (hrs)\end{tabular} & Test MSE & \begin{tabular}[c]{@{}l@{}}Runtime \\ (hrs)\end{tabular} \\ \hline
\multicolumn{1}{|l|}{None} & 1.8032 & 13:10 & 3.9742 & 22:55 \\
\multicolumn{1}{|l|}{Subsampling} & 0.1750 & 9:06 & 0.1005 & 18:44 \\
\multicolumn{1}{|l|}{Truncation} & 0.14713 & 9:28 & 0.6525 & 18:21 \\
\multicolumn{1}{|l|}{Both} & 0.3405 & 6:18 & 0.7672 & 13:11 \\ \hline
\end{tabular}%
}
\caption{Comparison of augmentation methods.}
\label{tab:aug}
\end{table}

The augmentation methods accelerate training, and achieve faster convergence. In Table \ref{tab:aug}, we report the generalization error calculated using test trajectories, and the wall time for training to complete 350 / 400 epochs, which both decrease. We report parameters used to run the data augmentation experiments.

The Sine Wave data set was generated using amplitudes and frequencies sampled from $(-8, 8)$ and $(2, 4)$ respectively. Phase was randomly sampled. $20000$ training trajectories of length 5 and with $200$ samples were generated. $2000$ validation trajectories were generated, and $2000$ test trajectories were generated. The Lotka-Volterra data set was generated using coefficients sampled from range $(0.5, 1.5), (0.5, 1.5), (1.5, 2.5), (0.5, 1.5)$ for $\alpha, \beta, \delta, \gamma$. The initial populations were sampled from range $(1.5, 2.5), (0.5, 1.5)$. $10000$ training trajectories were generated, with $500$ validation trajectories and $500$ test trajectories. Training trajectories contained $200$ samples, while validation and test trajectories contained $100$ samples. Trajectories were all of length $15$.

The Latent ODE in both Lotka Volterra and Sine Wave training runs were trained using Adamax with constant 1e-3 learning rate. A batch size of 256 was used. A latent dimension of 8 was used, while the encoder hidden state was of dimension 200. GRUs contained 200 units. A decoder network with 2 layer and 50 units with ReLU activation was used. In the Lotka Volterra data set, a 3 layer 200 units NN with Tanh activations parameterized both the encoder and latent Neural ODEs. In the Sine wave data set, the neural network width was reduced to 100.

\section{Character Trajectory Experimental Hyperparameters}
The LatSegODE base model was trained using a latent dimension of 8, and an encoder hidden state dimension of 16. Encoder GRUs contained 200 units. We parameterized the Neural ODEs in the encoder and latent dynamics using a 5 layer 200 unit NN with Tanh activations. The decoder network was a 3 layer 200 unit NN with ReLU activations. Hyperparameter search was performed by modifying the number of units in hidden layers and number of layers in all Neural ODE NNs. We started with 128 units per layer and 2 layers, and gradually increased the number of parameters until a good reconstruction was found.

The base model was trained using Adamax at a learning rate of 1e-3, reduced to 1e-4 when validation loss plateaued. Data augmentation was used, randomly sampling truncation bounds from (30, trajectory length) and the number of points to sub-sample from (30, trajectory length). KL annealing was used, such that the training run started with KL weight of 0, and reached a weight of 1 at epoch 50. We clipped gradients to a norm of 2. A fixed variance of 0.01 is used to compute the ELBO. Latent dynamics were solved using the dopri5 solver, with relative and absolute tolerances of 1e-5 and 1e-5. The encoder Neural ODE dynamics were solved using Euler's method.

At segmentation time, the LatSegODE used a fixed variance of 0.01 to compute marginal likelihood using 200 MC samples. A $K$ term of 100 was used. Time of observation was rounded to 2 decimal places. We set the minimum possible segment length to 20, similar to baselines.

\section{Sine Wave Ablation Study}
We report the effects of the number of training trajectories and number of samples per trajectory on LatSegODE performance. A unique LatSegODE is trained for each combination of the two parameters. We measure segmentation performance using the previously established metrics. The Sine Wave dataset is used with the same data generation and model hyper-parameters as in the main body Sine Wave experiments. The test set contains 75 trajectories each with zero to two changepoints. Each test trajectory contains 100 observed samples. The results are shown in Table \ref{tab:sine_ablate}. See Table \ref{tab:sine} for baseline results.

\begin{table}[h]
\centering
\resizebox{0.8\textwidth}{!}{%
\begin{tabular}{l|l|l|l|l|l|l|l|l|l|}
\cline{2-10}
 & \multicolumn{9}{l|}{\# Trajectory Samples} \\ \cline{2-10} 
 & \multicolumn{3}{l|}{50} & \multicolumn{3}{l|}{100} & \multicolumn{3}{l|}{200} \\ \hline
\multicolumn{1}{|l|}{\begin{tabular}[c]{@{}l@{}}\# Train \\ Trajectory\end{tabular}} & \begin{tabular}[c]{@{}l@{}}Rand \\ Index\end{tabular} & F1 Score & \begin{tabular}[c]{@{}l@{}}Hausdorff\\ Metric\end{tabular} & \begin{tabular}[c]{@{}l@{}}Rand \\ Index\end{tabular} & F1 Score & \begin{tabular}[c]{@{}l@{}}Hausdorff\\ Metric\end{tabular} & \begin{tabular}[c]{@{}l@{}}Rand \\ Index\end{tabular} & F1 Score & \begin{tabular}[c]{@{}l@{}}Hausdorff\\ Metric\end{tabular} \\ \hline
\multicolumn{1}{|l|}{3000} & 0.700 & 0.577 & 56.0 & 0.71 & 0.666 & 50.21 & 0.853 & 0.512 & 29.28 \\ \hline
\multicolumn{1}{|l|}{10000} & 0.663 & 0.601 & 49.6 & 0.842 & 0.813 & 28.2 & 0.968 & 0.961 & 6.29 \\ \hline
\multicolumn{1}{|l|}{30000} & 0.866 & 0.809 & 25.28 & 0.911 & 0.900 & 15.23 & 0.981 & 0.979 & 4.92 \\ \hline
\end{tabular}
}
\caption{LatSegODE performance measured by segmentation metrics, as a function of training set size and number of samples in training trajectories.}
\label{tab:sine_ablate}
\end{table}

\bibliography{main}
\bibliographystyle{style/icml2021}